\documentclass[10pt, a4paper]{article}
\usepackage{hybridreport}

\usepackage{array}
\usepackage{multirow}
\usepackage{wrapfig}
\usepackage{adjustbox}
\usepackage{placeins}
\usepackage[round]{natbib}

\graphicspath{{./}}

\newcommand{\pizero}{\ensuremath{\pi_0}}
\newcommand{\pihalf}{\ensuremath{\pi_{0.5}}}

\newcommand\blfootnote[1]{%
  \begingroup\renewcommand\thefootnote{}\footnote{#1}\addtocounter{footnote}{-1}\endgroup}

\reporttype{Technical Report}
\logo{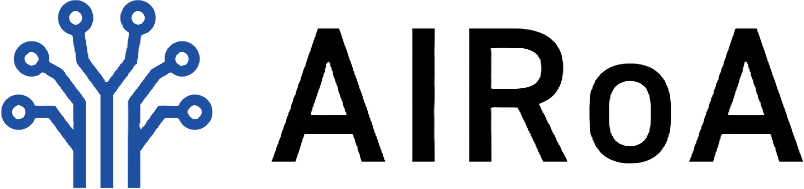}
\title{Continuous Reasoning for Vision-Language-Action}
\shorttitle{Continuous Reasoning for Vision-Language-Action}
\author{%
  Yueh-Hua (Kris) Wu\textsuperscript{*}\quad
  Tatsuya Matsushima\quad
  Kei Ota%
}
\institution{AIRoA}
\date{May 2026}
\correspondence{}

\abstracttext{%
  Natural language is a powerful reasoning medium for language and
  vision-language models, but it is mismatched to the granularity of continuous
  control. Text and explicit subgoals operate at task-level granularity, whereas
  vision-language-action (VLA) policies must choose actions at a much finer
  temporal scale; a single reasoning step can therefore span many action chunks
  while remaining only weakly coupled to the action needed now. This suggests a
  different question for VLA: what should play the role of language? We argue
  that a useful VLA reasoning medium must be shareable across model instances,
  verifiable through downstream action improvement, and aligned with temporally
  extended control structure. Based on this view, we propose \emph{Continuous
  Reasoning for Vision-Language-Action}. Our model first predicts continuous
  reasoning in the form of a structured set of continuous thoughts, then reuses
  them as shared context for chunk-structured action generation. We instantiate
  continuous reasoning as a shared Gaussian latent interface and train it with a
  self-verification objective in which an exponential-moving-average teacher must
  successfully consume the student's reasoning when predicting target actions.
  Empirically, Continuous Reasoning improves robustness on LIBERO-PRO and
  performs strongly on both HSR and TX-G2 (an AgiBot G2-compatible variant),
  raising mean real-robot subtask success over \pihalf{} by 40.4\% on TX-G2 and
  26.3\% on HSR. These results suggest that reasoning in VLA is less about
  producing extra tokens and more about learning a shareable and verifiable
  internal language for action.
}

\begin{document}
\maketitle
\blfootnote{\textsuperscript{*}Corresponding author: \href{mailto:kris.wu@airoa.org}{\texttt{kris.wu@airoa.org}}.}

\begin{center}
  \includegraphics[width=0.85\textwidth]{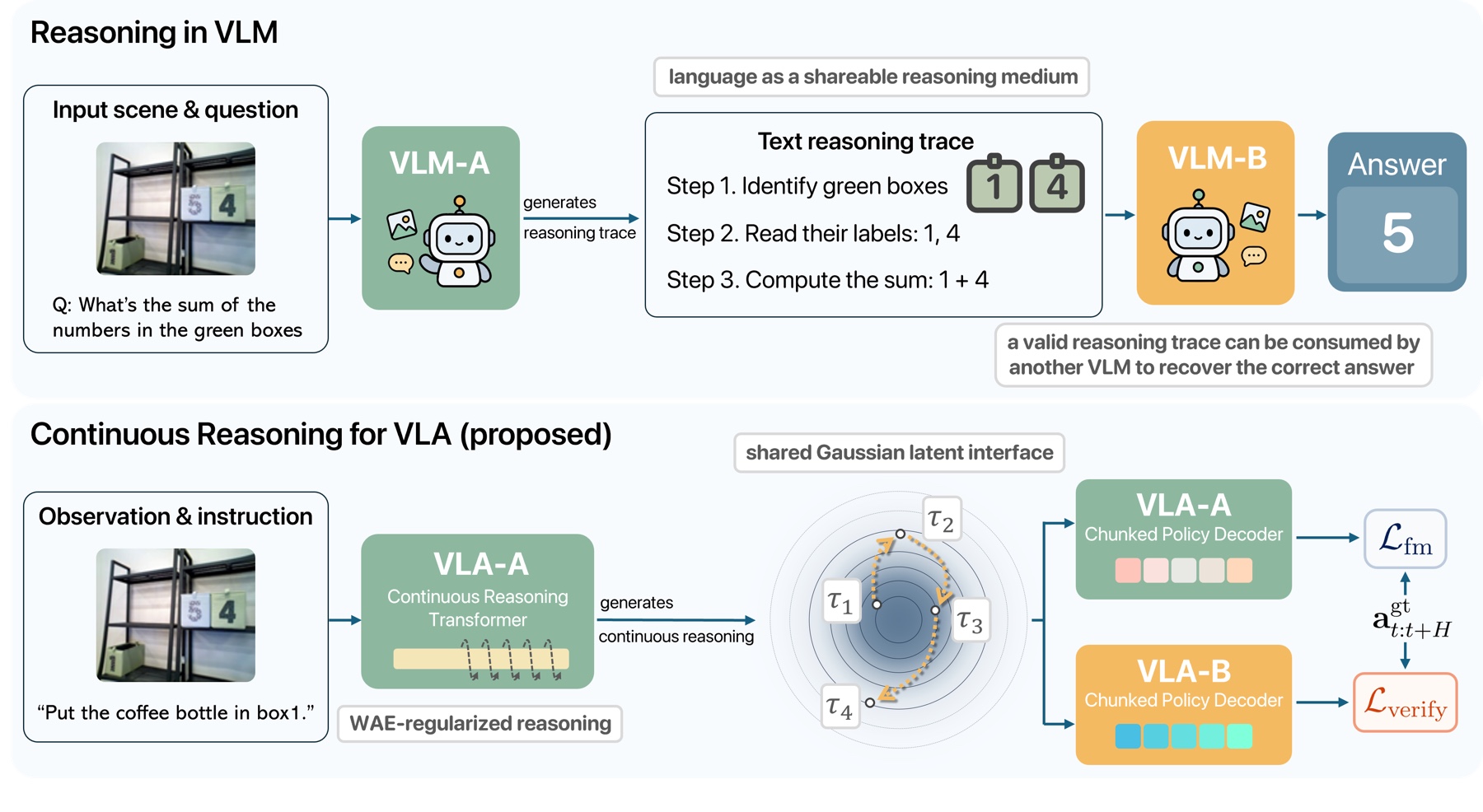}
\end{center}
\vspace{-0.6em}
{\small\color{TextMuted}%
  Top: in LLM/VLMs, language makes reasoning shareable and verifiable across
  model instances. Bottom: Continuous Reasoning seeks the analogous property for
  VLA through a shared Gaussian thought interface that another VLA instance must
  consume to predict better actions.\par}

\vspace{0.3em}
{\small\sffamily\color{TextMuted}\textbf{Keywords:}~Vision-language-action models, Reasoning\par}

\section{Introduction}

Vision-language-action (VLA) models have rapidly expanded the scope of robot policy learning by combining large-scale visual and language pretraining with action generation for manipulation and control. Recent systems have shown that internet-scale or multi-robot pretraining can substantially improve generalization, instruction following, and data efficiency in embodied decision making~\citep{rt1,rt2,palmE,octo,openvla,pi0}. At the same time, a parallel line of work has argued that successful embodied agents should not rely only on direct observation-to-action mapping, but should also perform some form of intermediate reasoning, planning, or decomposition before acting~\citep{saycan,innerMonologue,codeAsPolicies,voxposer}. As VLA policies become stronger and more general, the central question is no longer whether reasoning matters, but what form that reasoning should take for continuous control.

Natural language offers a useful source of inspiration for this question. In language and vision-language models, reasoning is effective not only because language is interpretable, but because it provides a shared medium: a reasoning trace can be read, reused, and built upon by another person or another model. Good intermediate reasoning is therefore not merely internally helpful; it is reusable across instances and externally verifiable. This suggests a more precise question for embodied control: \emph{what should play the role of language for VLA models?} If VLA systems are to reason before acting, they may also need a medium that preserves these functional advantages of language while remaining appropriate for the much finer granularity of continuous control.

Recent reasoning-oriented VLA methods implicitly propose several candidates for this medium, including textual or multimodal chain-of-thought traces~\citep{cotVla,halo,dualCotVla,dVla,deepThinkVla}, latent planning and compact reasoning states~\citep{thinkact,fastThinkAct,latentReasoningVla,last0,molmoact}, action-level reasoning abstractions~\citep{acotVla}, and interactive reasoning with external spatial guidance~\citep{guideThinkAct}. Each captures part of what makes language effective, but they typically emphasize explicit traces, latent planning efficiency, or external guidance rather than a reasoning interface that is simultaneously shareable, verifiable, and aligned with control granularity. Textual traces are naturally reusable and shareable, yet they remain mismatched to the temporal scale of control. A reasoning step such as ``first complete subtask 1, then subtask 2'' may be meaningful at the task level, but each such step typically unfolds over many action chunks before the goal is reached. The same reasoning unit must therefore support many successive action outputs, making its contribution to the action needed at the current control instant weak and indirect. Language is also a poor carrier for the precise geometric and dynamical structure required by action prediction. Visual subgoals and explicit foresight can expose future structure, but they often specify where the policy should eventually go rather than which action chunk should be produced now, while also incurring significant prediction cost. Latent reasoning moves closer to our setting by internalizing intermediate computation in continuous space. But improved action prediction alone does not certify good reasoning. This concern already appears in language-model reasoning, where recent work shows that outcome-based or RL-trained reasoning traces can improve final-answer accuracy without reliably yielding reasoning that is faithful, verifiable, or causally important~\citep{lanham2023faithfulness,mohammadi2025grpoFaithful,yu2026outcomeRewards}. In such cases, the added reasoning channel can help optimize behavior on seen problems without producing a process that is robust enough to generalize. The VLA setting inherits the same difficulty, but without a naturally shared medium like text; and text itself is mismatched to the granularity of continuous control. The right ``language'' for VLA must therefore do more than appear as an intermediate variable: it must match the structure of control.

In this paper, we therefore argue that the ``language'' of VLA should be modeled as a \emph{continuous internal interface} with three properties. First, it should be \emph{reusable}: if a reasoning trace is genuinely good, another model instance should be able to benefit from it rather than only the model that produced it. Second, it should be \emph{shareable}: the representation should live in a common latent space that can be transmitted and consumed across model instances, rather than existing only as an opaque byproduct of one forward pass. Third, it should be \emph{abstraction-aligned}: the representation should match the temporal granularity at which actions are organized, above low-level motor fluctuations but below free-form semantic language, so that it remains close enough to chunked action generation to guide control rather than merely describe it. Under this view, the goal of reasoning is not to generate extra text, but to form reusable symbols for future action generation. This also motivates our use of chunk-level action prediction as a natural unit for high-level control reasoning.

Based on this view, we propose \emph{Continuous Reasoning for Vision-Language-Action}. Our model first predicts continuous reasoning as a structured set of continuous thoughts, then reuses those thoughts as a shared reasoning context for chunk-structured action generation. To reduce arbitrariness in the thought space, we regularize the thoughts with a Wasserstein autoencoder (WAE) into a shared latent geometry~\citep{wae}. To test whether the learned thoughts are genuinely reusable, we introduce a self-verification objective in which an exponential-moving-average (EMA) teacher must successfully consume the student's thoughts in order to predict target actions~\citep{meanTeacher}. This self-verification mechanism makes the reasoning interface operational rather than decorative: a thought is only valuable if another model instance can use it to act better. We further align reasoning with high-level control structure through chunked generation, horizon-aware verification, and training choices that require action prediction to depend on the learned thought interface.

In summary, this paper makes three contributions. First, we formulate \emph{continuous reasoning} as a distinct interface for VLA, defined by reusability, shareability, and abstraction alignment. Second, we instantiate this view with a WAE-structured Gaussian latent interface, a self-verification objective, and chunk-aligned action generation. Third, we provide empirical evidence that these ingredients matter jointly, including mean real-robot subtask-success gains over \pihalf{} of 40.4\% on TX-G2, an AgiBot G2-compatible variant, and 26.3\% on HSR, while sharpening what reasoning should mean for continuous robot control.

\section{Related Work}
\paragraph{Reasoning in VLA.}
Recent VLA research increasingly studies explicit intermediate reasoning rather than pure observation-to-action mapping. Existing approaches span language-like or multimodal chain-of-thought traces~\citep{cotVla,halo,dualCotVla,dVla,deepThinkVla}, latent planning and compact reasoning states~\citep{thinkact,fastThinkAct,latentReasoningVla,last0,molmoact}, action-level reasoning abstractions~\citep{acotVla}, and interactive reasoning with external spatial guidance~\citep{guideThinkAct}. Related open-world and instruction-oriented systems also seek to preserve broader pretrained reasoning capabilities inside embodied models~\citep{palmE,chatVla2,instructVla,spatialVla}. Our work is closest to latent and action-oriented reasoning approaches, but differs in treating the intermediate representation as a shared interface whose usefulness is tested by another model instance and whose structure is aligned with chunk-level control.

\paragraph{Backbones, action generation, and planning.}
General-purpose VLA models such as RT-1, RT-2, PaLM-E, Octo, OpenVLA, OpenVLA-OFT, \pizero, and \pihalf establish the large-scale embodied policy-learning setting in which we operate~\citep{rt1,rt2,palmE,octo,openvla,openvlaOft,pi0,pi05}. In parallel, robot policy learning has developed structured action-generation and planning paradigms, including chunked action prediction, behavior tokenization, diffusion-based control, language-guided decomposition, and latent subgoal structure~\citep{act,behaviorTransformer,vqbet,diffusionPolicy,diffusionTransformerPolicy,baku,saycan,innerMonologue,codeAsPolicies,voxposer,subgoalDiffuser,foam,goalVla,playLmp}. Our focus is not primarily on scaling or decoding efficiency, but on what kind of reasoning interface should sit on top of a modern generative action policy.

\paragraph{Self-verification and latent structuring.}
Our training design is also related to teacher-student learning and structured latent modeling. Exponential-moving-average teachers are widely used to stabilize targets and encourage consistency across views or networks~\citep{meanTeacher,byol,dino}, while Wasserstein autoencoders impose a more organized latent geometry than unconstrained hidden states~\citep{wae}. Here their role is different: the EMA teacher tests whether predicted thoughts can be reused by another model instance, and WAE regularization supports a common latent space through which those thoughts can be shared.

\section{Continuous Reasoning}
We instantiate continuous reasoning as a shared internal interface between perception-language context and future action generation. Given an observation-instruction pair $(o, x)$, the model first generates a small set of raw continuous thoughts over dedicated reasoning slots, maps them into shared Gaussian latent codes, and then decodes them into a reasoning interface that is reused for chunk-structured action generation under a flow-matching policy. This design separates global reasoning from local generation: reasoning is produced once per horizon before action decoding and then reused across the full rollout, rather than being consumed through a rigid one-thought-per-chunk mapping at generation time.

\begin{figure}[t]
\centering
\includegraphics[width=\textwidth]{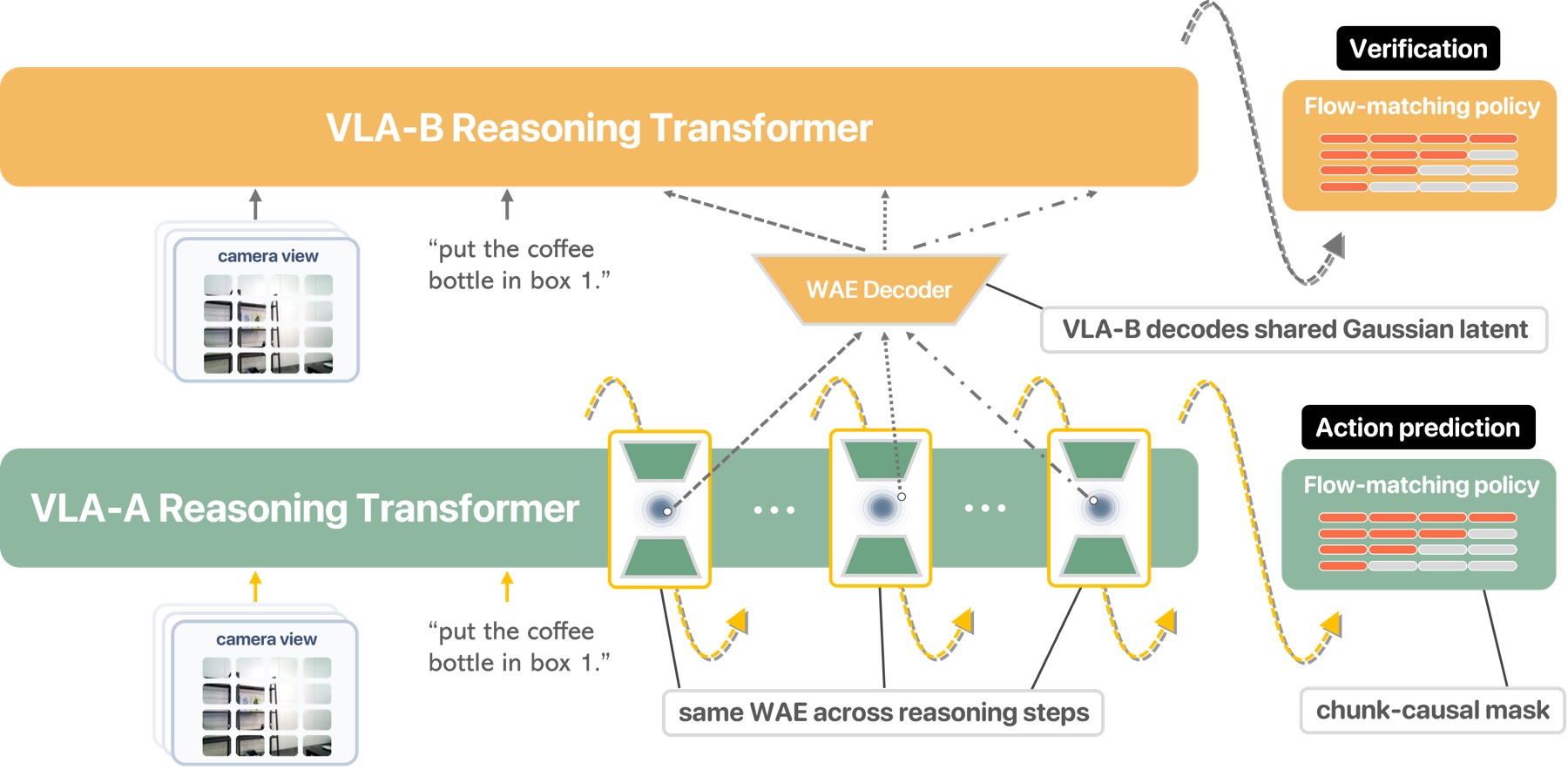}
\vspace{-1.2em}
\caption{Continuous Reasoning architecture. VLA-A generates continuous thoughts, maps them into shared Gaussian latent codes, and decodes them for chunk-causal action prediction. VLA-B decodes the same latent codes for verification; in our implementation, VLA-B is the EMA teacher.}
\label{fig:cr-architecture}
\vspace{-1.5em}
\end{figure}

Figure~\ref{fig:cr-architecture} summarizes the full architecture. The central path generates raw thoughts slot by slot, regularizes them into shared Gaussian latent codes, and reuses the student-decoded reasoning interface for chunk-structured action generation. During training, an EMA teacher receives the same latent codes and decodes them with its own WAE decoder in order to verify that the shared interface is reusable across model instances.

\paragraph{Problem setup and backbone.}
We consider language-conditioned action prediction from observations $o$, instruction $x$, and future action sequence $a \in \mathbb{R}^{H \times d}$ using a chunked flow-matching backbone. We divide the action horizon into $K$ temporally extended chunks, so that $H = K C$ for chunk size $C$. This chunked decomposition is important in our formulation: it defines the abstraction level at which reasoning should operate, above individual low-level action fluctuations but below free-form semantic language.

\paragraph{Instantiating continuous reasoning as structured thoughts.}
We represent continuous reasoning with $N_{\tau}$ raw thought vectors $\tau = [\tau_1, \ldots, \tau_{N_{\tau}}]$, where each $\tau_i \in \mathbb{R}^{D}$ and $N_{\tau}$ is independent of the action-chunk count $K$. We refer to $\tau$ as the raw continuous thoughts. The thought slots are generated sequentially rather than jointly: each slot is computed from the current prefix state, its decoded interface is written back into the prefix, and that decoded state is committed to the KV cache so that later slots can attend to earlier ones. These thoughts are not decoded into text and are not assigned one-to-one to local action chunks. Instead, they provide a compact horizon-level reasoning scaffold that is reused globally during action generation.

\paragraph{Shareable latent geometry.}
To make this reasoning medium shareable across model instances, we regularize the raw thoughts with a WAE~\citep{wae}. An encoder maps the thoughts to shared latent codes via $z = q_{\phi}(\tau)$, and a decoder produces the student-side reasoning interface $\hat{\tau}^{S} = p_{\psi}(z)$.
The EMA teacher receives the same shared latent codes $z$ but decodes them with its own WAE decoder, with EMA decoder parameters $\bar{\psi}$, yielding a teacher-side interface $\hat{\tau}^{\mathrm{ema}} = p_{\bar{\psi}}(z)$ before predicting actions. The WAE objective combines reconstruction and prior matching,
\[
\mathcal{L}_{\mathrm{wae}} =
\lambda_{\mathrm{rec}}\|\tau - \hat{\tau}^{S}\|_2^2 +
\lambda_{\mathrm{mmd}}\mathrm{MMD}(z, \mathcal{N}(0,I)).
\]
The purpose of this step is not to guarantee semantics by itself, but to impose a common latent geometry so that reasoning codes are less private and easier to transfer, decode, and reuse across model instances.

\paragraph{Flow-matching action prediction.}
Action prediction is trained with a flow-matching objective over noisy trajectories~\citep{pi05}. Given ground-truth action sequence $a$, Gaussian noise $\epsilon \sim \mathcal{N}(0, I)$, and interpolation time $t \in (0,1]$, we form the noisy action $x_t = (1-t)a + t\epsilon$ and target vector field $u_t = \epsilon - a$.
The student action model predicts a vector field $v_{\theta}(o, x, x_t, t; \hat{\tau}^{S})$ conditioned on the student-decoded reasoning interface, and is trained with
\[
\mathcal{L}_{\mathrm{fm}} = \mathbb{E}\left[\|v_{\theta}(o, x, x_t, t; \hat{\tau}^{S}) - u_t\|_2^2\right].
\]
This decoder sits between fully bidirectional flow-matching action prediction and step-wise autoregressive decoding. Within each chunk, the flow-matching head predicts a temporally extended continuous action segment without left-to-right decoding over individual motor steps, following the broader line of chunked generative action policies~\citep{act,diffusionPolicy,pi0,openvlaOft}. Across chunks, we use a block-causal attention mask: action tokens in the same chunk share one attention block, while later chunks can attend to earlier chunks. Thus temporal dependency is modeled at the level of high-level action chunks rather than microscopic control fluctuations, and the same decoded reasoning interface remains available throughout action generation.

\paragraph{Reusable reasoning through self-verification.}
To encourage the reasoning interface to be reusable rather than merely helpful to the producing model, we introduce a self-verification objective using an EMA teacher~\citep{meanTeacher}. Let $\bar{\theta}$ denote the EMA parameters, and let $\hat{\tau}^{\mathrm{ema}}$ denote the teacher-decoded interface obtained from the student's shared latent codes. During training, the teacher must decode the same latent codes and use the resulting interface to predict the same target action field:
\[
\mathcal{L}_{\mathrm{verify}}
=
\mathbb{E}\left[\|v_{\bar{\theta}}(o, x, x_t, t; \hat{\tau}^{\mathrm{ema}}) - u_t\|_2^2\right].
\]
This objective operationalizes reusability at training time: another model instance must successfully decode the same shared latent codes and use the resulting interface when predicting target actions.

\paragraph{Abstraction alignment and interface dependence.}
The final requirement is that the reasoning medium capture the right level of abstraction for control. We enforce this in two ways. First, the chunk-level decoder places temporal dependency at the level of action chunks rather than individual motor steps: within each chunk, flow matching predicts a temporally extended action segment in parallel, while across chunks a block-causal mask allows later chunks to attend to earlier ones. Second, in the main configuration used throughout this paper, we do not use separate per-chunk conditioning paths that would let local action prediction bypass the shared reasoning interface, so action generation must consume the decoded reasoning prefix itself rather than a private shortcut. These choices keep continuous reasoning close to chunk-level control while discouraging it from collapsing into low-level motor perturbations.

\paragraph{Overall objective.}
The final training objective combines action prediction, latent structuring, and self-verification:
\[
\mathcal{L}
=
\mathcal{L}_{\mathrm{fm}}
+
\mathcal{L}_{\mathrm{wae}}
+
\lambda_{\mathrm{verify}}\mathcal{L}_{\mathrm{verify}}.
\]
Overall, the model first predicts raw continuous thoughts $\tau$, maps them into shared Gaussian latent codes $z$, and then asks both the student policy and the EMA teacher to act from decoded interfaces derived from the same latent codes. The training objective therefore ties action prediction to a reasoning medium that is shared across model instances and aligned with chunk-level control, rather than to a producer-specific latent shortcut. We refer to this overall framework as \emph{continuous reasoning} for VLA.

\section{Experiments}
We begin with LIBERO-PRO~\citep{liberoPro}, a robustness-oriented benchmark that is substantially more challenging than standard LIBERO evaluation because success depends not only on in-distribution task execution, but also on maintaining performance under controlled distribution shifts. Rather than measuring a single aggregate success rate, LIBERO-PRO exposes policies to multiple perturbation families across several LIBERO suites, making it possible to separate semantic robustness from spatial robustness and task-level generalization. This makes it a better testbed for our claim than standard LIBERO success rates, which are now often saturated by modern VLA systems.

\begin{wrapfigure}[14]{r}{0.39\textwidth}
\vspace{-0.5em}
\centering
\includegraphics[width=\linewidth]{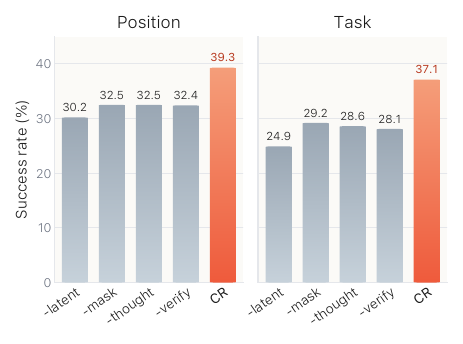}
\vspace{-2.3em}
\caption{LIBERO-PRO ablations. Labels indicate the removed component.}
\label{fig:libero-ablation}
\end{wrapfigure}

\begin{table}[!t]
\centering
\scriptsize
\setlength{\tabcolsep}{2.7pt}
\caption{LIBERO-PRO perturbation breakdown across all four LIBERO suites.}
\label{tab:libero-pro-eval}
\renewcommand{\arraystretch}{1.18}
\adjustbox{max width=\textwidth}{%
\begin{tabular}{>{\raggedright\arraybackslash}m{3.6cm}|*{4}{>{\centering\arraybackslash}m{0.68cm}}|*{4}{>{\centering\arraybackslash}m{0.68cm}}|*{4}{>{\centering\arraybackslash}m{0.68cm}}|*{4}{>{\centering\arraybackslash}m{0.68cm}}}
\toprule
\multirow{2}{3.6cm}{\textbf{Methods}} & \multicolumn{4}{c|}{\texttt{libero\_10}} & \multicolumn{4}{c|}{\texttt{libero\_goal}} & \multicolumn{4}{c|}{\texttt{libero\_object}} & \multicolumn{4}{c}{\texttt{libero\_spatial}} \\
& Obj. & Pos. & Sem. & Task & Obj. & Pos. & Sem. & Task & Obj. & Pos. & Sem. & Task & Obj. & Pos. & Sem. & Task \\
\midrule
OpenVLA-OFT~\citep{openvlaOft} & 5.5 & 0.0 & 38.5 & 0.0 & 8.5 & 0.0 & 43.5 & 2.5 & 66.5 & 9.5 & 89.0 & 0.0 & 30.0 & 13.5 & 65.0 & 0.0 \\
X-VLA~\citep{xVla} & 61.0 & 1.0 & 70.5 & 19.5 & 71.5 & 1.0 & 94.0 & 7.5 & 91.5 & 4.5 & 98.0 & 0.0 & 89.5 & 0.0 & 69.0 & 38.5 \\
VLA-Adapter~\citep{vlaAdapter} & 47.0 & 0.0 & \textbf{91.0} & 10.0 & 61.0 & 0.0 & 75.0 & 12.0 & 89.0 & 0.0 & \textbf{99.0} & 8.0 & 98.0 & 0.0 & \textbf{98.0} & 49.0 \\
\pihalf{}~\citep{pi05} & 66.0 & 6.0 & \textbf{91.0} & 17.0 & \textbf{90.0} & 29.0 & 95.0 & 17.0 & 89.0 & 19.0 & 95.0 & 10.0 & \textbf{99.0} & \textbf{53.0} & 97.0 & 55.0 \\
\rowcolor{TableHighlight}
\textbf{CR (Ours)} & \textbf{66.5} & \textbf{15.5} & 88.5 & \textbf{28.5} & 81.5 & \textbf{36.0} & \textbf{96.0} & \textbf{33.0} & \textbf{93.0} & \textbf{54.5} & \textbf{99.0} & \textbf{27.0} & 97.5 & 51.0 & 96.0 & \textbf{60.0} \\
\bottomrule
\end{tabular}
}
\vspace{-1.5em}
\end{table}

Table~\ref{tab:libero-pro-eval} reports the main simulation comparison against strong VLA baselines. We report four official perturbation types used throughout this paper: \emph{object}, which changes object appearance such as color, texture, or size; \emph{position}, which changes object placement; \emph{semantic}, which paraphrases the task instruction; and \emph{task}, which evaluates transfer to different target task variants. Among these, \emph{position} and \emph{task} are the most revealing because they require the policy to re-anchor action under a changed layout or pursue a modified goal, rather than merely tolerate appearance shifts or instruction paraphrases.

Continuous Reasoning yields the strongest overall robustness profile under the current LIBERO-PRO protocol, achieving the best suite mean on all four suites and improving the average suite mean from 58.0 with \pihalf{} to 64.0. These gains come primarily from the two perturbations that most directly test action retargeting: averaged across suites, position success rises from 26.8 to 39.3 and task success from 24.8 to 37.1, while object and semantic performance remain competitive. In contrast, object and semantic perturbations can be spuriously forgiving under overfit action templates, because a policy may still obtain high scores by replaying nearly the same action sequence while largely ignoring the changed appearance or wording. The more revealing failure mode is therefore re-anchoring action under shifted spatial layouts or goal structure, and Continuous Reasoning improves exactly this regime.

\begin{figure}[!t]
\centering
\includegraphics[width=\textwidth]{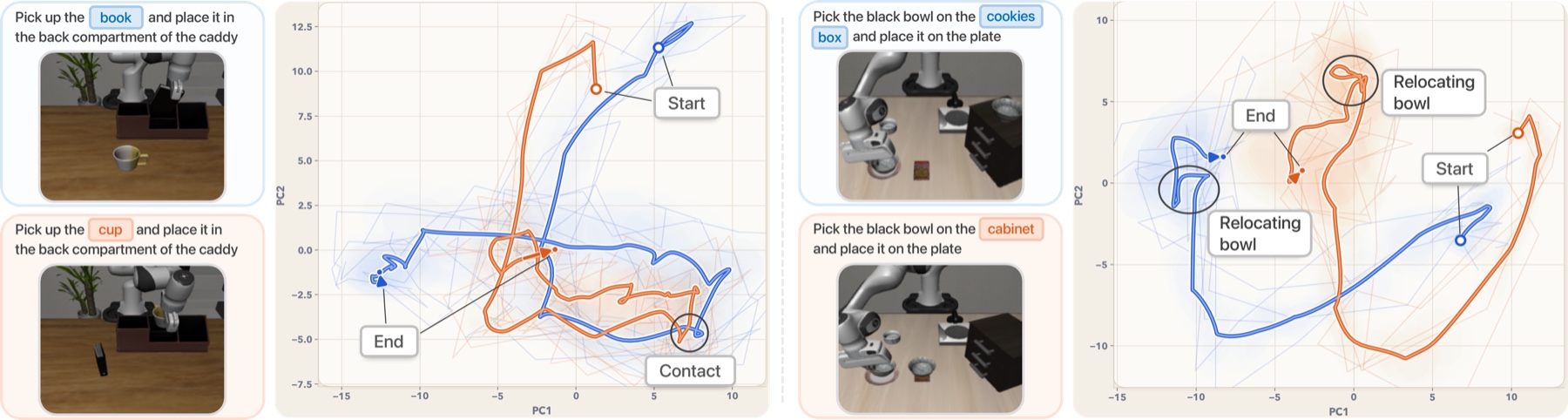}
\vspace{-0.5em}
\caption{PCA visualization of continuous reasoning trajectories on paired LIBERO-PRO scenes. Each instruction is evaluated with five rollouts from the same initial scene, so differences in the projected trajectories reflect instruction-dependent reasoning under matched scene context.}
\label{fig:cr-vis}
\vspace{-0.8em}
\end{figure}

Figure~\ref{fig:cr-vis} provides a diagnostic view of the learned continuous reasoning interface beyond scalar success rates. We deliberately compare instruction pairs from the same initial scene and collect five rollouts for each instruction so that differences in the projected trajectories reflect task-dependent reasoning rather than scene variation. On the left, both tasks require reaching toward nearby objects and placing them into the same caddy compartment, so the early approach curves are similar without collapsing into an identical template. The trajectories become especially close near contact, where grasp geometry is similar for the cup and the book, and then separate again during placement because the larger book requires a different insertion strategy. On the right, the two tasks relocate different bowls to the same target plate. Here the pre-grasp portions differ substantially because the target bowls begin far apart, forcing different approach trajectories and reasoning before contact, while the post-grasp relocation phase exhibits a visibly related pattern across both instructions. This shared relocation structure appears specifically after the bowl has been grasped, suggesting that Continuous Reasoning is not merely encoding scene identity, but reorganizing around task phase and object-specific control demands.
Additional paired-scene latent visualizations on LIBERO and TX-G2 are provided in Appendix~\ref{sec:appendix-libero-ablation} and Appendix~\ref{sec:appendix-real-robot}.

\paragraph{Ablation study.}
To isolate where these gains come from, we ablate Gaussian latent structuring, chunk-causal action masking, explicit continuous thoughts, and self-verification while keeping the same \pihalf{} backbone and LIBERO-PRO evaluation protocol. Figure~\ref{fig:libero-ablation} focuses on the two perturbations that matter most for our claim, \emph{position} and \emph{task}, while the full table is provided in Appendix~\ref{sec:appendix-libero-ablation}. All four ablations reduce both \emph{position} and \emph{task} success, with the largest degradation appearing when Gaussian latent structuring is removed, whereas \emph{object} and \emph{semantic} remain comparatively stable. This again points to improved spatial re-anchoring and task-level adaptation rather than a generic regularization effect. We report CALVIN ABC$\rightarrow$D in Appendix~\ref{sec:appendix-calvin} as a complementary long-horizon benchmark, where Continuous Reasoning remains competitive, indicating that these LIBERO-PRO gains do not come at the expense of standard long-horizon execution.

\paragraph{Real-robot evaluation.}
\begin{wrapfigure}{r}{0.36\textwidth}
\vspace{-1.2em}
\centering
\includegraphics[width=\linewidth]{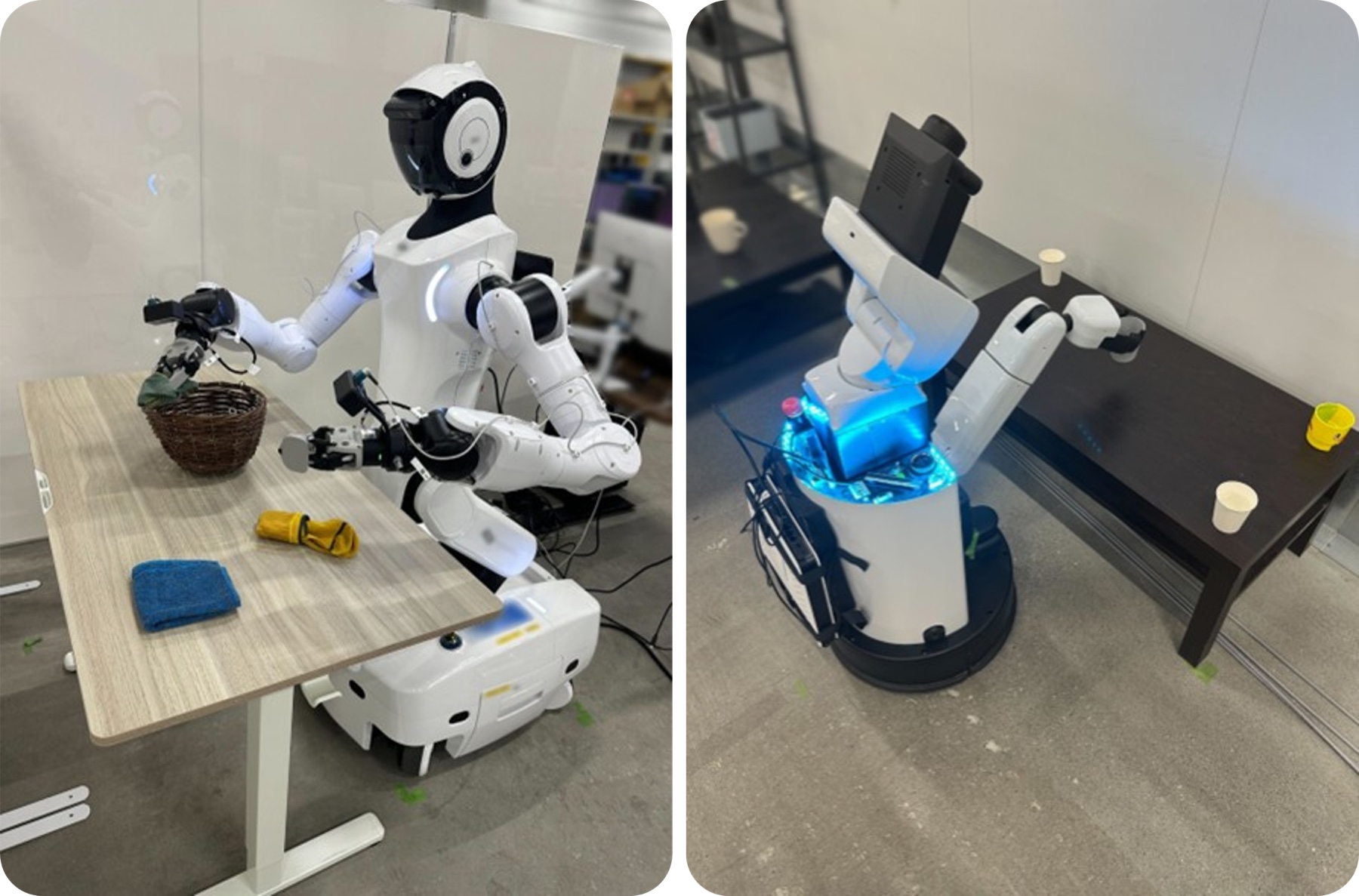}
\vspace{-1.6em}
\caption{TX-G2 and HSR.}
\label{fig:real-platforms}
\end{wrapfigure}
We additionally evaluate Continuous Reasoning on two real-robot platforms, HSR and TX-G2 (an AgiBot G2-compatible variant). Figure~\ref{fig:real-platforms} shows the two platforms used in our experiments. All reported real-robot baselines are fine-tuned for 150k steps. We exclude OpenVLA-OFT from this comparison because our RTX 5070 evaluation setup runs out of memory on both platforms under the deployed real-robot inference pipeline. In the main text, we report \emph{Subtask} success, the mean success rate over subtasks within each task, and defer full E2E results, task descriptions, and protocol details to Appendix~\ref{sec:appendix-real-robot}.

\paragraph{TX-G2.}

TX-G2 is a bimanual platform with three cameras, and we query the deployed policy at 10\,Hz. We evaluate each task with ten episodes, of which eight use in-distribution placements and two use substantially shifted out-of-distribution placements. For fairness, all methods are evaluated on the same ten initial states and object placements for each task. Objects may appear on either the left or right side of the workspace, so the policy must also decide which arm to use. We consider four long-horizon tasks: \emph{Cutlery Transfer}, \emph{Bowl Stacking}, \emph{Clothes Sorting}, and \emph{Dish Racking}.

\begin{table}[t]
\centering
\scriptsize
\setlength{\tabcolsep}{3.0pt}
\renewcommand{\arraystretch}{1.16}
\newcommand{\stdcell}[2]{\mbox{#1 {\color[gray]{0.45}\fontsize{5}{5}\selectfont$\pm$#2}}}
\caption{Real-robot subtask success on TX-G2 and HSR. Main-text results report mean subtask success with standard deviation; full E2E results are deferred to Appendix~\ref{sec:appendix-real-robot}.}
\label{tab:real-robot-subtask}
\vspace{-0.3em}
\adjustbox{max width=\textwidth}{%
\begin{tabular}{>{\raggedright\arraybackslash}m{1.65cm}|*{4}{>{\centering\arraybackslash}m{1.02cm}}|*{4}{>{\centering\arraybackslash}m{1.00cm}}}
\toprule
\multirow{2}{1.65cm}{\textbf{Method}} & \multicolumn{4}{c|}{\textbf{TX-G2}} & \multicolumn{4}{c}{\textbf{HSR}} \\
& \textbf{Cutl.} & \textbf{Bowl} & \textbf{Cloth.} & \textbf{Dish} & \textbf{Bott.-1} & \textbf{Bott.-2} & \textbf{Box} & \textbf{Mug} \\
\midrule
\mbox{X-VLA} & \stdcell{0.0}{0.0} & \stdcell{7.5}{4.0} & \stdcell{5.0}{2.6} & \stdcell{5.0}{3.4} & \stdcell{8.3}{7.6} & \stdcell{4.2}{3.8} & \stdcell{0.0}{0.0} & \stdcell{25.0}{12.3} \\
\mbox{VLA-Adapter} & \stdcell{0.0}{0.0} & \stdcell{0.0}{0.0} & \stdcell{0.0}{0.0} & \stdcell{0.0}{0.0} & \stdcell{0.0}{0.0} & \stdcell{0.0}{0.0} & \stdcell{0.0}{0.0} & \stdcell{0.0}{0.0} \\
\pihalf{} & \stdcell{5.0}{3.2} & \stdcell{47.5}{6.8} & \stdcell{83.3}{4.8} & \stdcell{60.0}{7.2} & \stdcell{\textbf{83.3}}{9.6} & \stdcell{45.8}{8.5} & \stdcell{41.7}{14.0} & \stdcell{66.7}{9.6} \\
\rowcolor{TableHighlight}
CR (Ours) & \stdcell{\textbf{22.5}}{6.4} & \stdcell{\textbf{70.0}}{7.2} & \stdcell{\textbf{95.0}}{2.7} & \stdcell{\textbf{87.5}}{4.5} & \stdcell{\textbf{83.3}}{10.8} & \stdcell{\textbf{83.3}}{6.8} & \stdcell{\textbf{58.3}}{12.3} & \stdcell{\textbf{75.0}}{10.2} \\
\bottomrule
\end{tabular}%
}
\vspace{-1.5em}
\end{table}
\global\let\stdcell\relax

Continuous Reasoning already shows a clear advantage over \pihalf{} on TX-G2, improving subtask success on all four tasks. The gains are especially large on \emph{Bowl Stacking} and \emph{Dish Racking}, where multi-stage object handling and precise placement both matter. Appendix~\ref{sec:appendix-real-robot} shows that VLA-Adapter's 0.0 performance is more consistent with a weakly task-conditioned policy than with a major deployment bug.

\begin{figure}[!t]
\centering
\includegraphics[width=\textwidth]{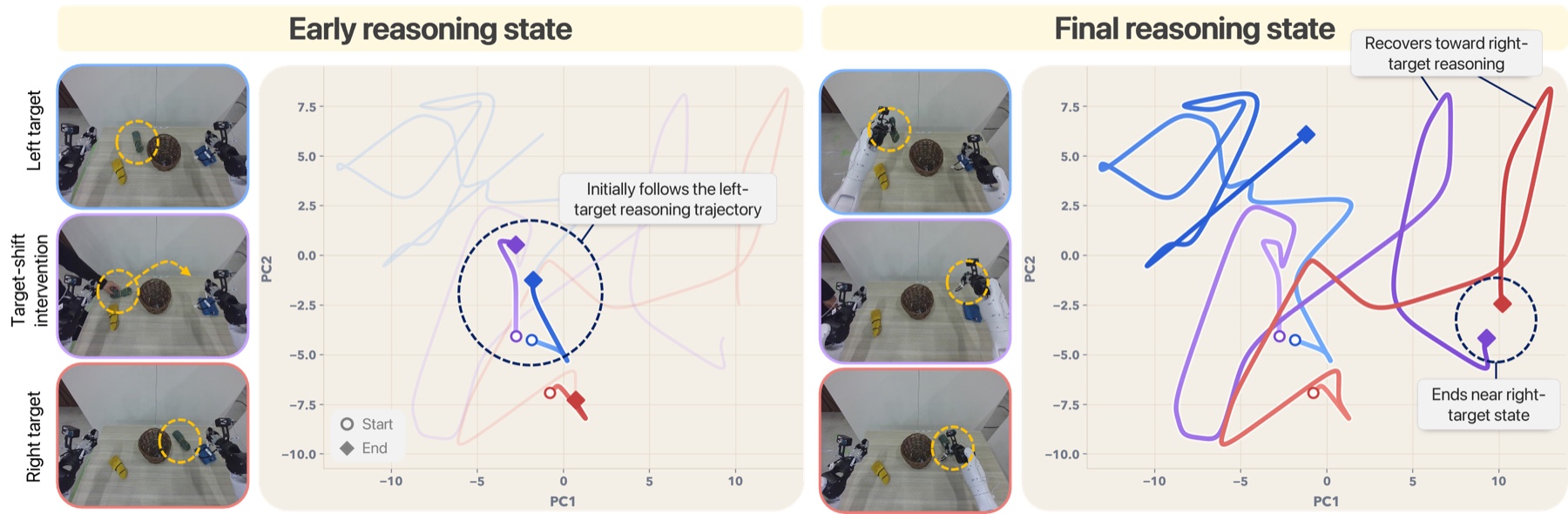}
\vspace{-0.5em}
\caption{PCA visualization of continuous reasoning on the TX-G2 subtask ``pick up the green socks.'' Top: the target pair of green socks starts on the left. Bottom: it starts on the right. Middle: it starts on the left and is thrown to the right mid-episode by an external perturbation.}
\label{fig:g2-reasoning}
\end{figure}

Figure~\ref{fig:g2-reasoning} provides a diagnostic view of continuous reasoning on the TX-G2 clothes-sorting subtask ``pick up the green socks.'' We visualize three matched scene variants: the target pair of green socks starts on the left (top), starts on the right (bottom), or starts on the left and is abruptly thrown to the right by an external perturbation during execution (middle). This kind of within-episode target displacement does not appear in the dataset. Nevertheless, the perturbed middle rollout initially follows the left-target reasoning pattern, but its final reasoning state moves toward the right-target configuration. This behavior is consistent with online re-anchoring of the reasoning interface after the target object has moved, rather than rigidly replaying the original plan.

\begin{figure}[!t]
\centering
\includegraphics[width=\textwidth]{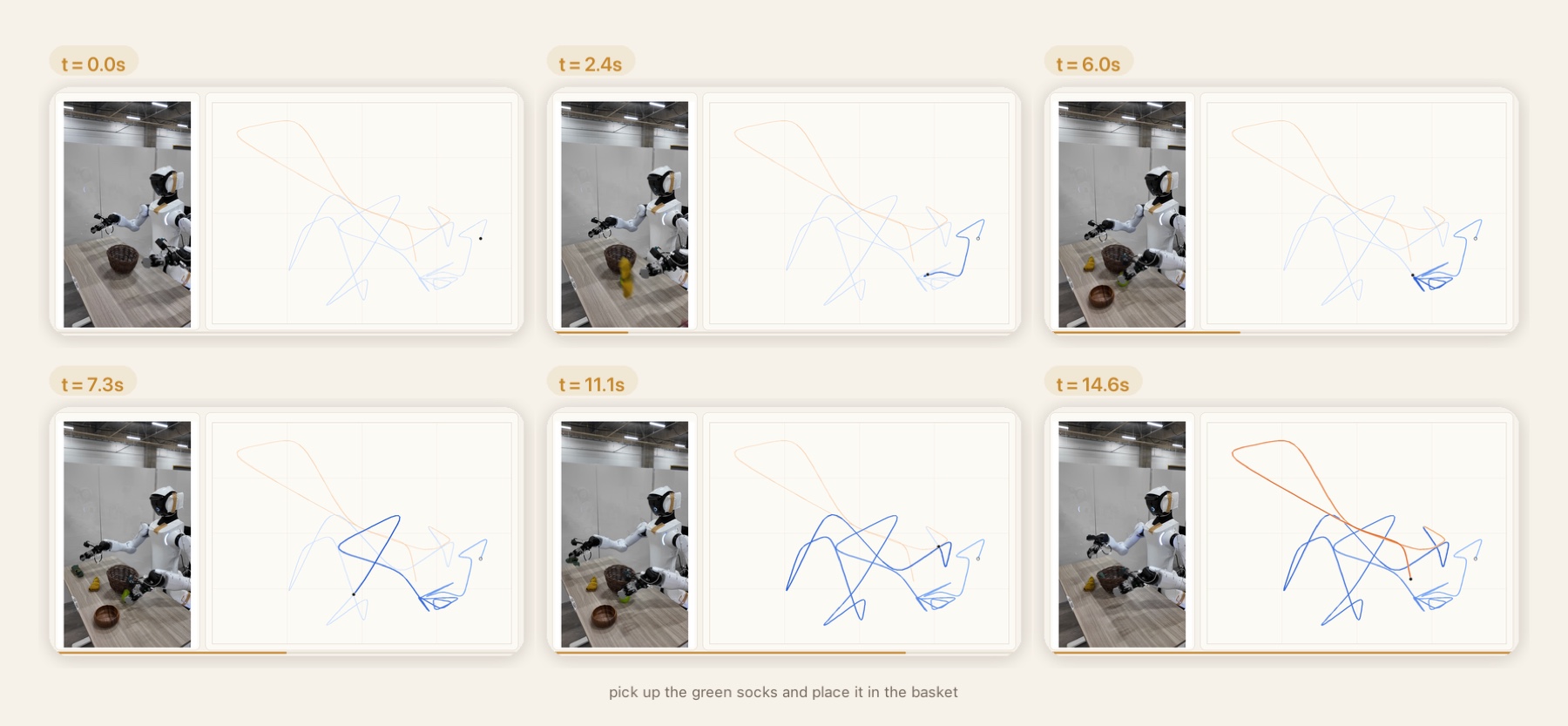}
\vspace{-0.5em}
\caption{TX-G2 latent visualization under dynamic object injection for the clothes-sorting task. Before the target green socks appear, the continuous reasoning latent remains clustered near the same state around $t{=}6.0$ despite incoming distractor objects. Once the green socks enter the scene, the latent sharply transitions into a different region and then proceeds through distinct pickup and placement phases. In the plot, blue markers denote the pickup stage and brown markers denote the placement stage, showing that the two phases require different latent reasoning states.}
\label{fig:g2-latent-injection}
\end{figure}

Figure~\ref{fig:g2-latent-injection} probes a more adversarial TX-G2 setup in which the workspace initially contains no relevant object and a human successively throws different objects into view. Only green socks, yellow socks, a blue handkerchief, and the basket appear in the training data; at test time, we additionally inject distractors such as green fruit. The resulting latent trajectories reveal whether Continuous Reasoning stays inert under irrelevant events and reconfigures only when the true target becomes available.

\paragraph{HSR.}
HSR emphasizes long-horizon mobile manipulation with locomotion, and we query the deployed policy at 2\,Hz. We use four tasks that combine navigation, manipulation, and constrained placement geometry: \emph{Coffee Bottle $\rightarrow$ Box}, \emph{Coffee Bottles $\rightarrow$ Table}, \emph{Box Rearrangement}, and \emph{Mug Rectangle}. Full task descriptions are deferred to Appendix~\ref{sec:appendix-real-robot}.

Continuous Reasoning also improves HSR subtask success on three of the four mobile-manipulation tasks, with the largest gain on \emph{Coffee Bottles $\rightarrow$ Table}. This task is the most demanding in the suite because it requires repeated locomotion and manipulation over a long action chain: approach the shelf, pick a bottle, carry it to the table, return to the shelf, and repeat for the second bottle.

\FloatBarrier
\section{Conclusions}
We presented Continuous Reasoning for Vision-Language-Action, a framework that treats reasoning in VLA not as an explicit language-like trace, but as a reusable continuous interface for future action generation. Our formulation learns structured continuous thoughts that are shared across model instances, organized in a common latent space, and reused as a reasoning context for chunk-structured action generation.

Our results support this view on both challenging generalization and real-robot evaluation. Continuous Reasoning improves robustness on LIBERO-PRO and performs strongly on both HSR and TX-G2, suggesting that reasoning for continuous control is better modeled as an internal language for action rather than as extra reasoning tokens. More broadly, we hope this work helps shift the discussion in VLA from whether models should reason before acting to what kind of reasoning interface they should learn.

\separator

\subsection*{Acknowledgments}
{\small\color{TextMuted}%
This paper is based on results obtained from a project, JPNP25015, commissioned
by the New Energy and Industrial Technology Development Organization (NEDO). We
thank Ryosuke Takanami and Haru Kondoh for their assistance in the real-robot
experiments.}

\bibliographystyle{corlabbrvnat}
\bibliography{references}

\clearpage
\appendix
\setcounter{table}{0}
\setcounter{figure}{0}
\renewcommand{\thetable}{A\arabic{table}}
\renewcommand{\thefigure}{A\arabic{figure}}
\renewcommand{\theHtable}{appendix.table.\arabic{table}}
\renewcommand{\theHfigure}{appendix.figure.\arabic{figure}}

\section{Experimental Configuration}

Unless otherwise noted, all experiments in this paper use the same default H16
training configuration summarized in Table~\ref{tab:h16-config}. This includes
the main LIBERO-PRO comparison, the LIBERO-PRO ablations, the CALVIN results,
and the real-robot experiments. The benchmark-specific sections below therefore
focus on evaluation protocol details rather than on different model settings.

\begin{table}[H]
\centering
\scriptsize
\setlength{\tabcolsep}{6.0pt}
\renewcommand{\arraystretch}{1.10}
\caption{Default H16 training configuration used across all experiments in this paper unless otherwise noted.}
\label{tab:h16-config}
\begin{tabular}{lll}
\toprule
\textbf{Symbol} & \textbf{Value} & \textbf{Setting} \\
\midrule
$H$ & $16$ & action horizon \\
$C$ & $4$ & action chunk length \\
$K$ & $4$ & number of chunks ($H = KC$) \\
$N_{\tau}$ & $2$ & maximum thought slots \\
$d_z$ & $128$ & thought latent dimension \\
$\lambda_{\mathrm{verify}}$ & $0.1$ & constant verification weight \\
$\gamma_{\mathrm{EMA}}$ & $0.994$ & EMA decay \\
$\mathcal{S}_{\mathrm{curr}}$ & $[0, 2000]$ & thought curriculum steps \\
\bottomrule
\end{tabular}
\end{table}
 
\paragraph{Distributed WAE-MMD implementation.}
In multi-GPU, multi-node training, the WAE prior-matching loss should not be
computed independently on local shards. Instead, we first aggregate the latent
codes and Gaussian prior samples across all devices to form one global batch,
and then compute a single IMQ-kernel MMD loss on that global batch. This detail
matters because MMD is a batch-level discrepancy: if each device computes its
own local MMD, the kernel only sees small per-device sub-batches and no longer
matches the intended global latent distribution. We use this same distributed
WAE computation across all experiments in this paper unless otherwise noted.

\section{LIBERO-PRO Evaluation Details}

Our LIBERO-PRO results follow the official static benchmark protocol released with LIBERO-PRO. We evaluate on four suites: \texttt{libero\_spatial}, \texttt{libero\_goal}, \texttt{libero\_10}, and \texttt{libero\_object}. Each suite contains 10 tasks. For every suite--perturbation pair, we run 20 trials per task, yielding 200 evaluation episodes for each cell reported in Table~\ref{tab:libero-pro-eval}.

We report four official perturbation types: \emph{object}, \emph{position}, \emph{semantic}, and \emph{task}. Success rates are reported as percentages over the corresponding 200 episodes. The \emph{Suite Mean} in Table~\ref{tab:libero-pro-eval} is the arithmetic mean of these four perturbation success rates within the same suite block.

This protocol is intentionally more diagnostic than a single in-distribution success rate. Because each perturbation family isolates a different source of shift, the full LIBERO-PRO table reveals whether a policy mainly benefits from appearance robustness, instruction robustness, spatial retargeting, or task-level generalization.

\section{LIBERO-PRO Ablation Details}
\label{sec:appendix-libero-ablation}

Table~\ref{tab:libero-ablation-full} reports the full LIBERO-PRO ablation breakdown averaged across all four suites. The variants correspond to removing one major component of Continuous Reasoning at a time while keeping the same \pihalf{} backbone and training budget.

\begin{table}[H]
\centering
\scriptsize
\setlength{\tabcolsep}{5.4pt}
\caption{Full LIBERO-PRO ablation results averaged across the four suites. The full Continuous Reasoning model is lightly highlighted for readability.}
\label{tab:libero-ablation-full}
\renewcommand{\arraystretch}{1.12}
\begin{tabular}{lccccc}
\toprule
\textbf{Method} & \textbf{Object} & \textbf{Position} & \textbf{Semantic} & \textbf{Task} & \textbf{Overall} \\
\midrule
\rowcolor{TableHighlight}
\textbf{Full CR} & 84.6 & \textbf{39.3} & 94.9 & \textbf{37.1} & \textbf{64.0} \\
w/o Gaussian latent space & 86.0 & 30.2 & 95.5 & 24.9 & 59.2 \\
w/o chunk-causal mask & \textbf{86.5} & 32.5 & 96.0 & 29.2 & 61.1 \\
w/o continuous thoughts & 84.9 & 32.5 & \textbf{96.5} & 28.6 & 60.6 \\
w/o self-verification & 86.0 & 32.4 & 95.5 & 28.1 & 60.5 \\
\bottomrule
\end{tabular}
\end{table}

For clarity, we summarize the precise implementation of each ablation below.
\paragraph{Ablation definitions.}
\textbf{w/o self-verification} removes the verification loss $L_{\mathrm{verify}}$ and does not train against the EMA teacher. The model still predicts continuous thoughts and still uses latent structuring, but those thoughts are no longer required to improve another model instance.

\textbf{w/o Gaussian latent space} removes the WAE objective and the Gaussian bottleneck. Instead of decoding thoughts from a structured latent code, we pass the model's original uncompressed continuous thought vectors directly to the action predictor and to the EMA teacher. This tests whether latent structuring itself matters beyond simply having a continuous intermediate representation.

\textbf{w/o chunk-causal mask} keeps continuous thoughts, latent structuring, and self-verification, but removes the block-causal dependency across action chunks. In this variant, the flow-matching decoder treats the full action horizon as a single block rather than imposing the AR-like temporal dependency used by the full method.

\textbf{w/o continuous thoughts} is the action-only variant. We remove the continuous reasoning branch entirely and keep only the chunk-structured flow-matching action decoder. Because no thought interface is produced, this variant also has no latent-structuring objective and no self-verification objective.

The same pattern as the main-text plot is visible here. Object and semantic perturbations remain relatively stable, but every ablation substantially reduces position and task success. This supports the claim that Continuous Reasoning primarily improves the components of control that require spatial retargeting and adaptation to changed task structure.

\paragraph{Additional LIBERO paired-scene latent diagnostics.}
We additionally visualize paired LIBERO scenes that share the same initial state but differ only in instruction, to further probe how the latent reasoning interface reorganizes under changed goals. These plots are diagnostic rather than mechanistic proofs, but they provide a more direct qualitative view of what the learned latent preserves and where it diverges.

\begin{figure}[H]
\centering
\begin{minipage}[t]{0.48\textwidth}
    \centering
    \includegraphics[width=\textwidth]{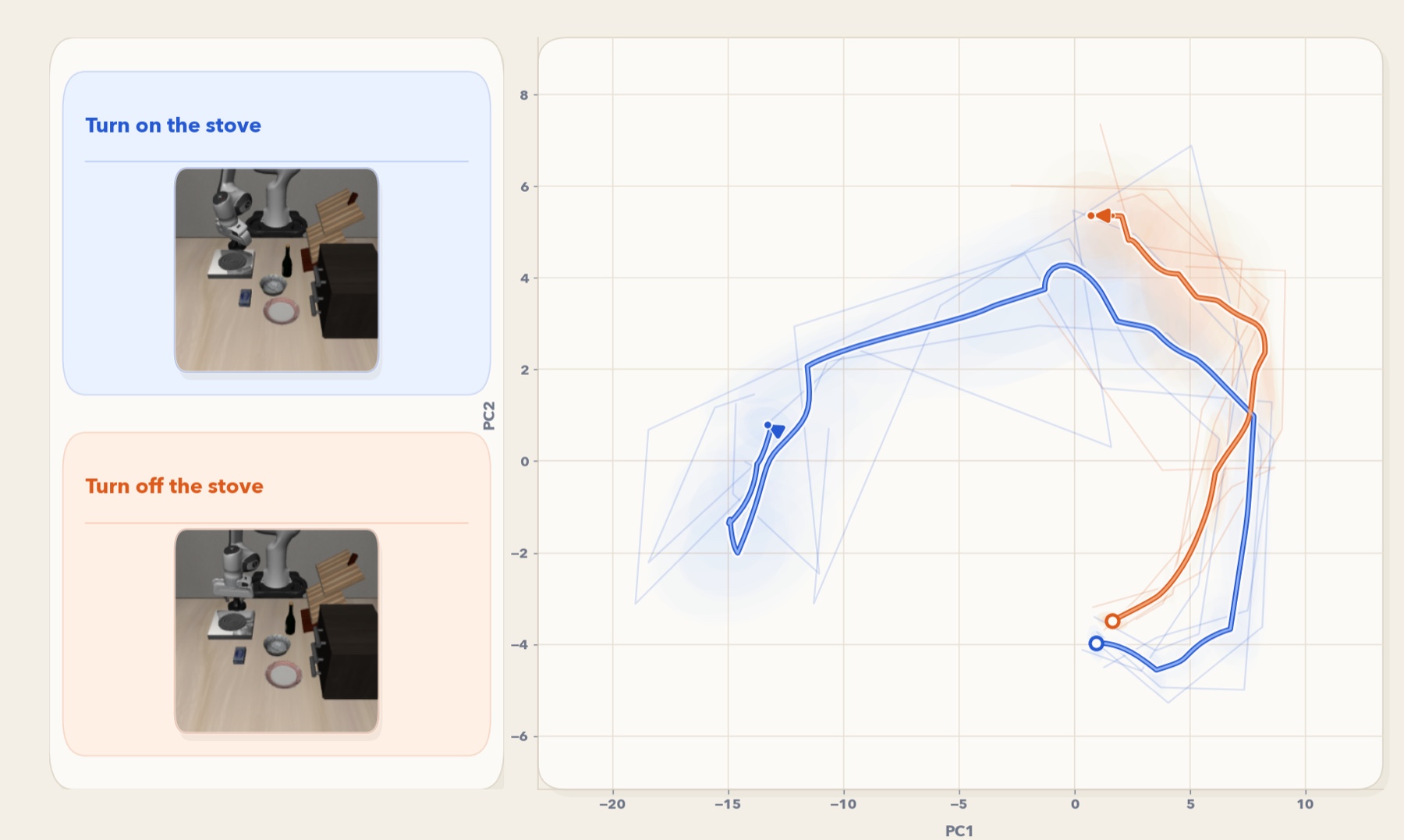}
    \par\vspace{0.2em}
    {\scriptsize \textbf{Turn on vs. turn off the stove}}
\end{minipage}\hfill
\begin{minipage}[t]{0.48\textwidth}
    \centering
    \includegraphics[width=\textwidth]{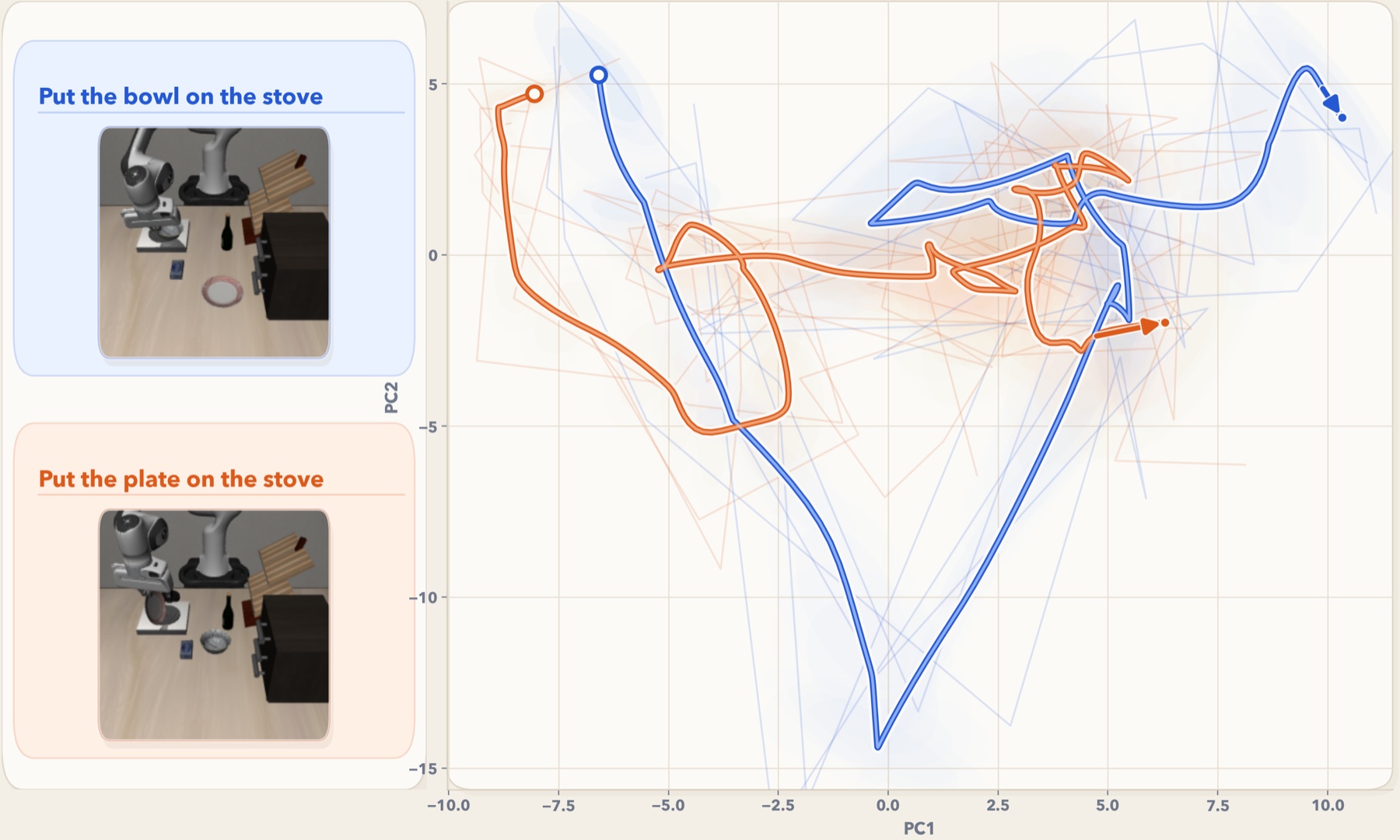}
    \par\vspace{0.2em}
    {\scriptsize \textbf{Place different objects onto the stove}}
\end{minipage}
\caption{Additional LIBERO paired-scene latent visualizations under matched initial states. Left: the two instructions initially produce closely aligned latent trajectories because both begin with the same approach-to-stove motion; the \emph{turn off} rollout then terminates early because the stove already starts in the off state. Right: when placing different objects onto the stove, the latent state is similar near the initial and final phases, where the start and placement geometry are analogous, but diverges in the middle while the policy reaches to different object locations.}
\label{fig:appendix-libero-latent-vis}
\end{figure}

\section{Additional Results on CALVIN}
\label{sec:appendix-calvin}

We report CALVIN ABC$\rightarrow$D as a complementary long-horizon benchmark. While it is not the primary benchmark used to support our main empirical claims, it provides a useful check that Continuous Reasoning remains competitive on standard long-horizon manipulation evaluation. We report only average sequence length here.

\begin{figure}[H]
\centering
\includegraphics[width=0.48\textwidth]{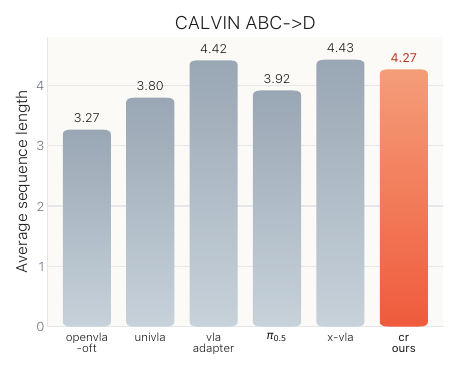}
\caption{CALVIN ABC$\rightarrow$D average sequence length.}
\label{fig:calvin-abcd}
\end{figure}

As shown in Figure~\ref{fig:calvin-abcd}, Continuous Reasoning remains close to the strongest baselines on CALVIN ABC$\rightarrow$D, indicating that the robustness gains observed in the main text are not purchased by sacrificing standard long-horizon performance.

\section{Real-Robot Evaluation Details}
\label{sec:appendix-real-robot}

\paragraph{Shared evaluation setup.}
All reported real-robot baselines are fine-tuned for 150k steps. We exclude OpenVLA-OFT from both TX-G2 and HSR because our RTX 5070 evaluation setup runs out of memory under the deployed real-robot inference pipeline.

\paragraph{TX-G2 protocol.}
TX-G2 (an AgiBot G2-compatible variant) is a bimanual platform equipped with three cameras. Each task is evaluated with 10 episodes: 8 use in-distribution object placements that remain close to the training distribution without exactly duplicating dataset states, and 2 use substantially shifted out-of-distribution placements. For fairness, all compared methods are evaluated on the same 10 initial states and object placements for each task. Objects may appear on either the left or right side of the workspace, so the policy must also decide which arm to use.

\paragraph{TX-G2 dataset composition.}
The TX-G2 finetuning dataset contains 1198 source trajectories across four long-horizon task families. Table~\ref{tab:g2-dataset-composition} summarizes the trajectory count for each family, and the short descriptions below give the canonical subtask sequence that appears most often in each case.

\begin{wraptable}[12]{r}{0.42\textwidth}
\vspace{-1.0em}
\centering
\scriptsize
\setlength{\tabcolsep}{5.0pt}
\renewcommand{\arraystretch}{1.12}
\caption{Composition of the TX-G2 finetuning dataset.}
\label{tab:g2-dataset-composition}
\begin{tabular}{lr}
\toprule
\textbf{Task family} & \textbf{Trajectories} \\
\midrule
Bowl Stacking & 129 \\
Clothes Sorting & 514 \\
Cutlery Transfer & 206 \\
Dish Racking & 349 \\
\bottomrule
\end{tabular}
\vspace{-1.0em}
\end{wraptable}

\paragraph{TX-G2 tasks.}
\textbf{Cutlery Transfer} requires moving a spoon and a fork from one bowl to another. The utensils are thin and difficult to grasp, and lifting the source bowl counts as failure.

\textbf{Bowl Stacking} presents three bowls with different colors and asks the robot to stack the instructed target bowls.

\textbf{Clothes Sorting} places two pairs of socks together with a handkerchief on the table and requires the instructed items to be placed into a basket.

\textbf{Dish Racking} uses two colored plates and requires placing the instructed plate onto a dish rack with limited clearance, making orientation correction particularly important.

\paragraph{Canonical TX-G2 subtask sequences.}
\textbf{Bowl Stacking.} A typical sequence is: pick up the yellow bowl from the desk, stack the yellow bowl on the grey bowl, pick up the light-blue bowl from the desk, and stack the light-blue bowl on the yellow bowl.

\textbf{Clothes Sorting.} A typical sequence is: pick up the green socks from the desk, place the green socks into the basket, pick up the handkerchief from the desk, place the handkerchief into the basket, pick up the yellow socks from the desk, and place the yellow socks into the basket.

\textbf{Cutlery Transfer.} A typical sequence is: pick up the light-blue spoon from the grey bowl, place the spoon in the yellow bowl, pick up the pink fork from the grey bowl, and place the fork in the yellow bowl.

\textbf{Dish Racking.} A typical sequence is: pick up the yellow dish from the desk, place the yellow dish in the wooden dish rack, pick up the green dish from the desk, and place the green dish in the wooden dish rack.

\begin{figure}[H]
\centering
\begin{minipage}[t]{0.24\textwidth}
    \centering
    \includegraphics[width=\textwidth]{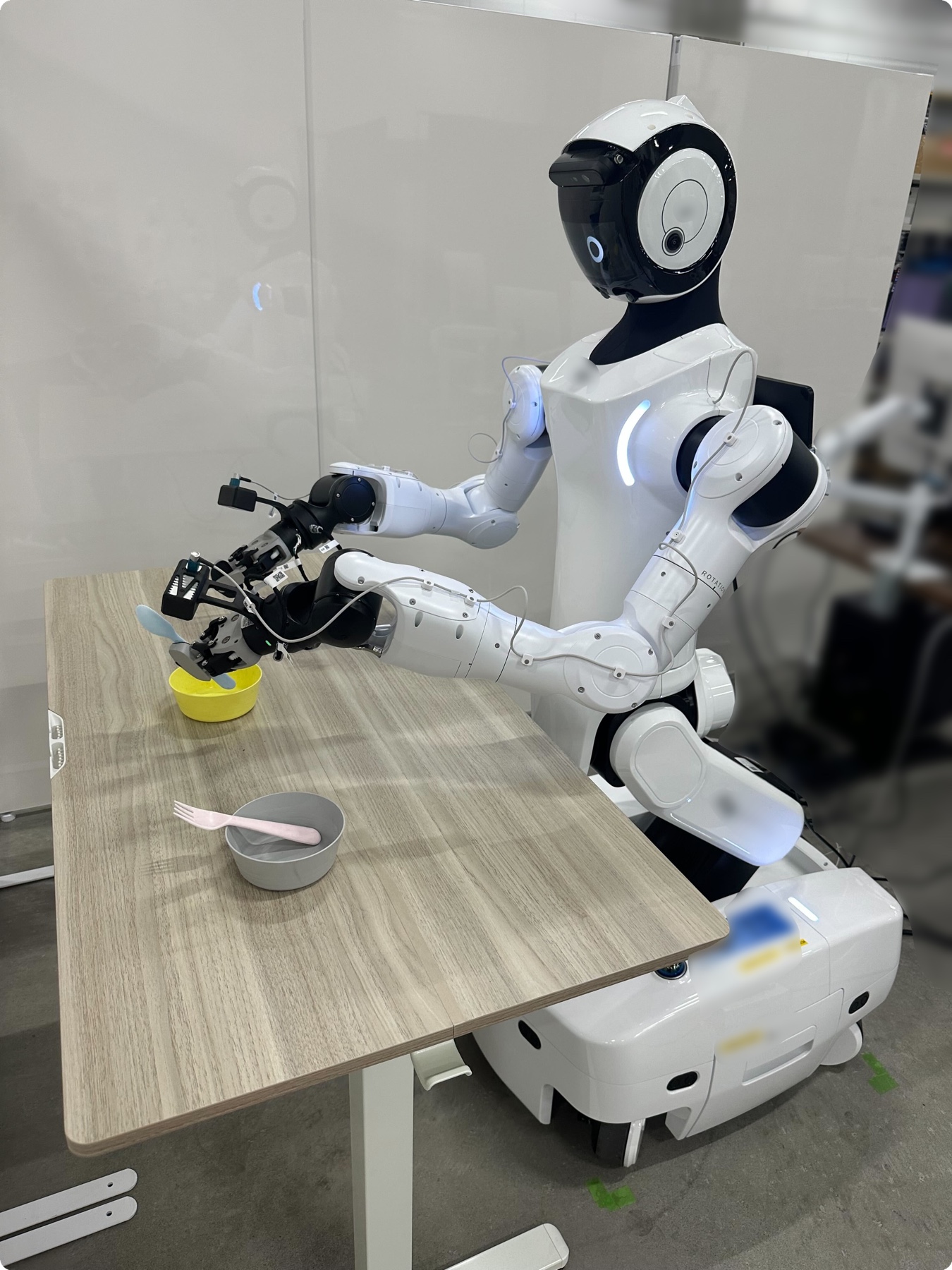}
    \par\vspace{0.2em}
    {\scriptsize \textbf{Cutlery Transfer}}
\end{minipage}\hfill
\begin{minipage}[t]{0.24\textwidth}
    \centering
    \includegraphics[width=\textwidth]{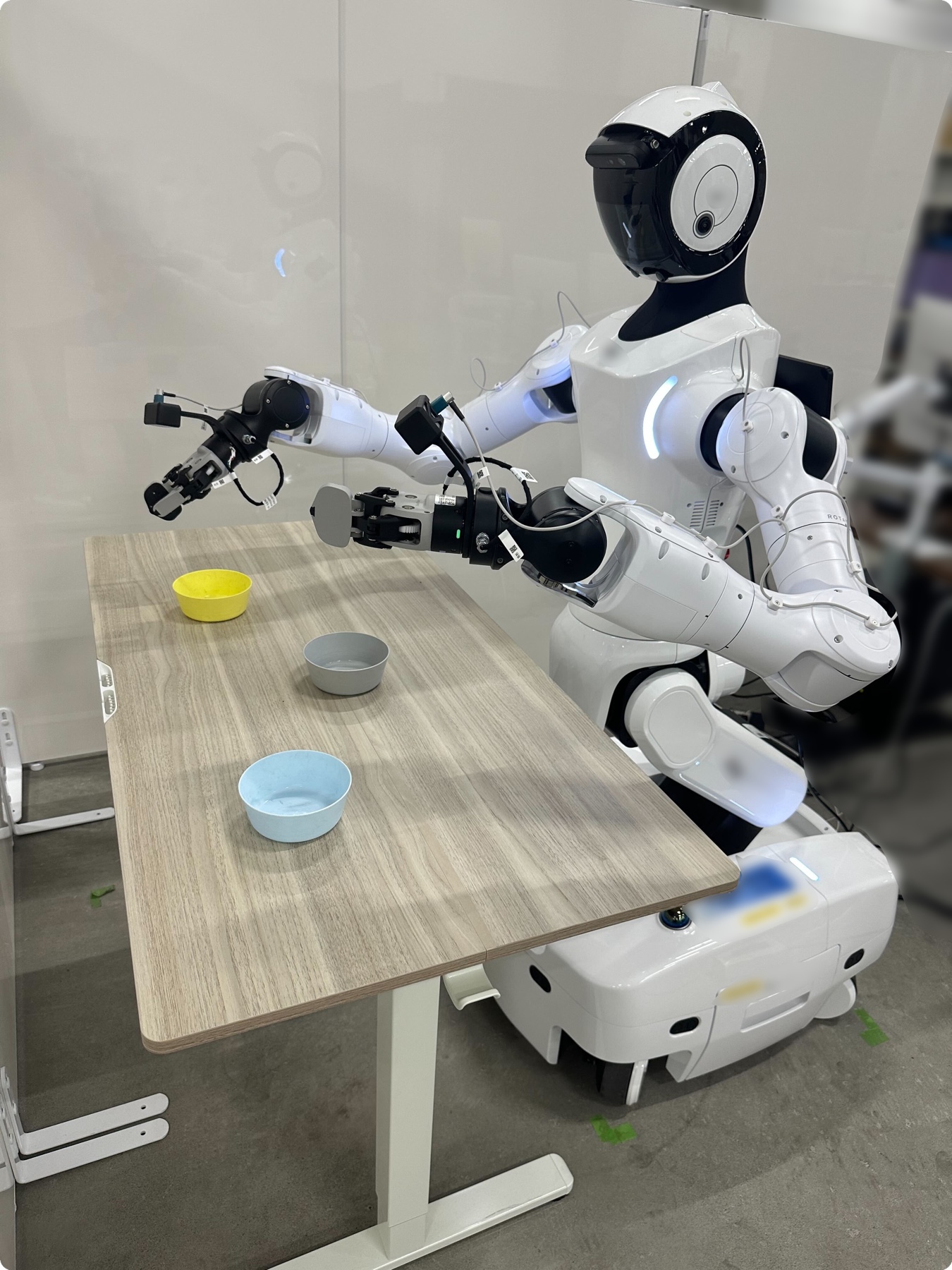}
    \par\vspace{0.2em}
    {\scriptsize \textbf{Bowl Stacking}}
\end{minipage}\hfill
\begin{minipage}[t]{0.24\textwidth}
    \centering
    \includegraphics[width=\textwidth]{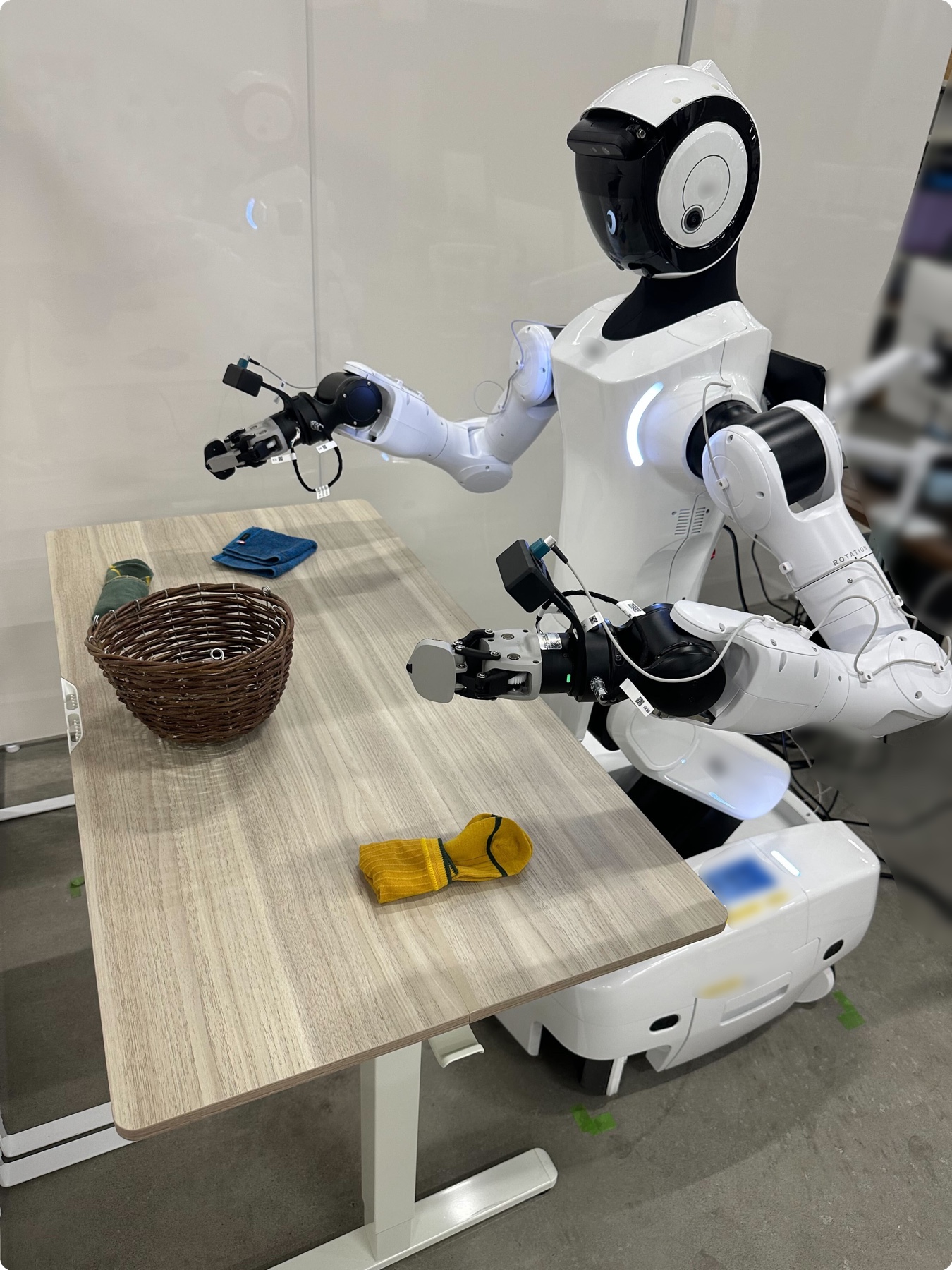}
    \par\vspace{0.2em}
    {\scriptsize \textbf{Clothes Sorting}}
\end{minipage}\hfill
\begin{minipage}[t]{0.24\textwidth}
    \centering
    \includegraphics[width=\textwidth]{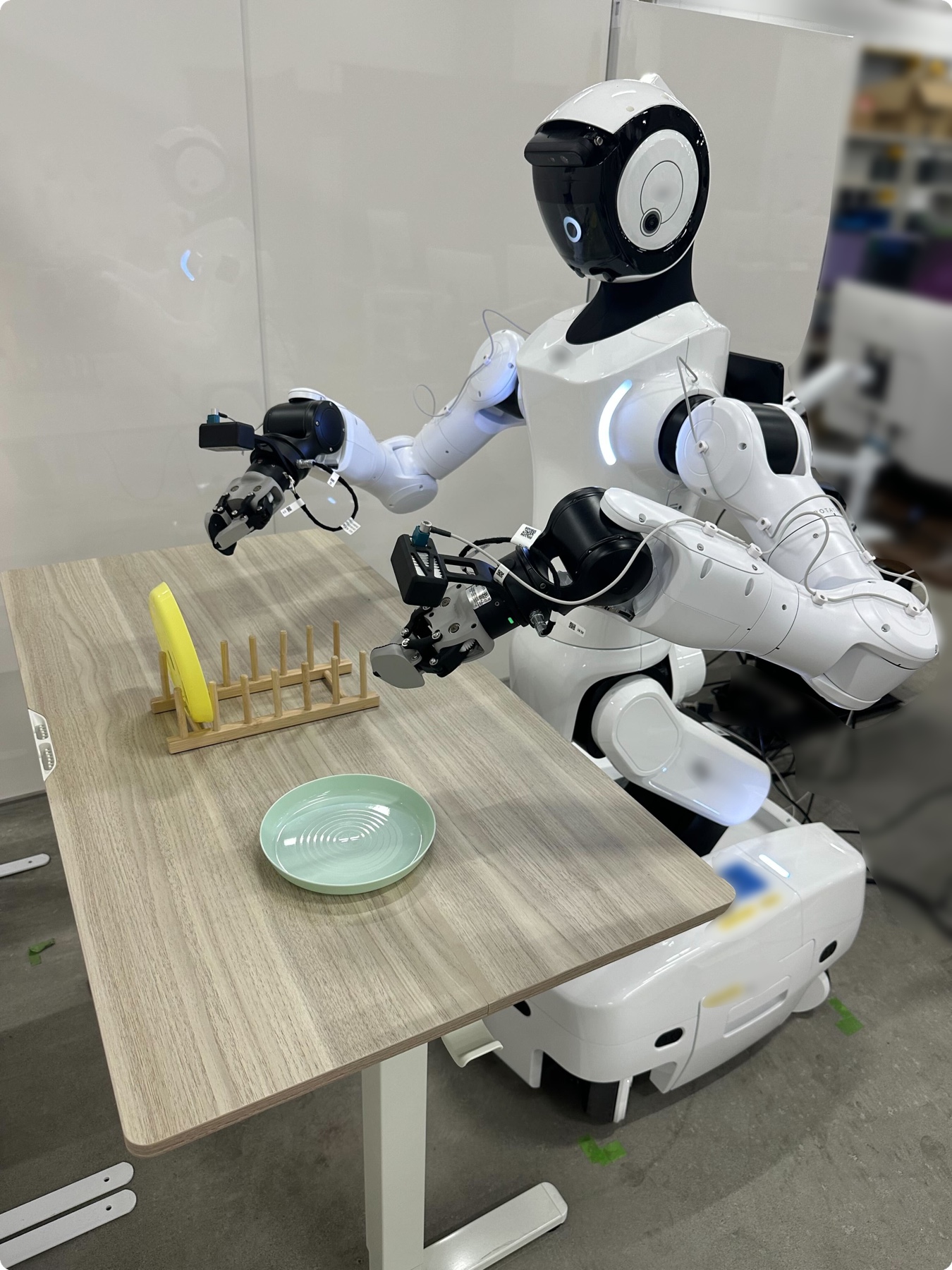}
    \par\vspace{0.2em}
    {\scriptsize \textbf{Dish Racking}}
\end{minipage}
\caption{Representative TX-G2 trajectory snapshots for the four long-horizon tasks described above. Each panel shows a rollout of the corresponding task family used in our real-robot evaluation.}
\label{fig:appendix-g2-trajectories}
\end{figure}

\paragraph{Additional TX-G2 closed-loop robustness under human intervention.}
We also evaluate a more challenging TX-G2 rollout in which the workspace initially starts empty and the policy must follow a strict ordered instruction sequence: pick up the green socks and place them into the basket, then the handkerchief, and finally the yellow socks. During execution, a nearby human continuously perturbs the scene by moving the basket and introducing additional objects. This setting is substantially harder than the standard clothes-sorting evaluation because the policy must both identify the correct target and preserve the required ordering under ongoing scene changes.

\begin{figure}[t]
\centering
\includegraphics[width=\textwidth]{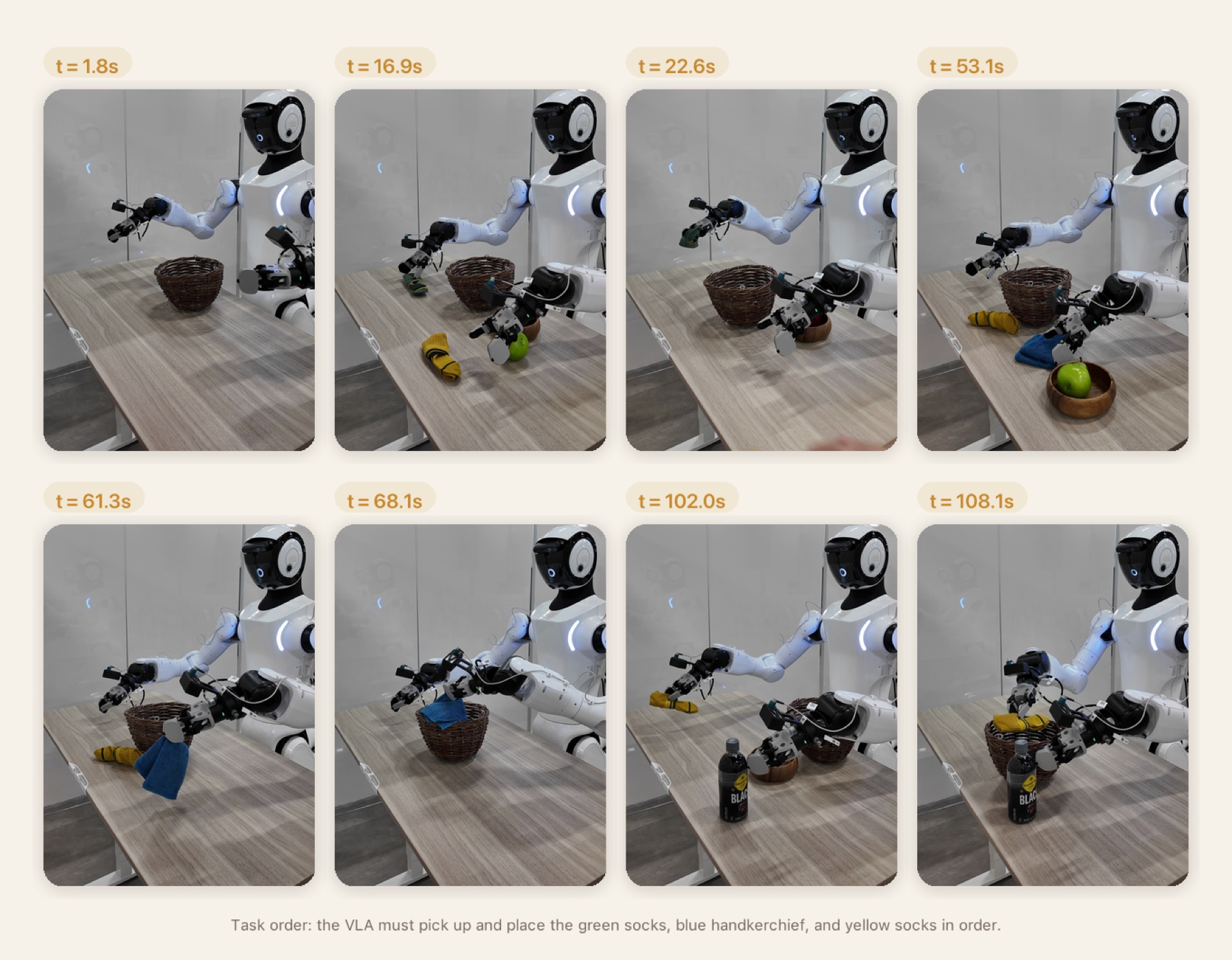}
\caption{TX-G2 closed-loop rollout under continuous human intervention. The table starts empty, and a nearby human repeatedly moves the basket and inserts additional objects while the policy must still follow the ordered sequence \emph{green socks $\rightarrow$ handkerchief $\rightarrow$ yellow socks}. Despite these perturbations, Continuous Reasoning stably completes the required sequence and places the instructed items into the basket in order.}
\label{fig:appendix-g2-ordered-rollout}
\end{figure}

\paragraph{HSR tasks.}
HSR emphasizes long-horizon mobile manipulation with locomotion.

\paragraph{HSR dataset composition.}
The HSR finetuning dataset contains 1205 source trajectories across four long-horizon mobile-manipulation task families. Table~\ref{tab:hsr-dataset-composition} summarizes the trajectory count for each family, and the short descriptions below give the canonical subtask sequence that appears most often in each case.

\begin{table}[H]
\centering
\scriptsize
\setlength{\tabcolsep}{5.0pt}
\renewcommand{\arraystretch}{1.12}
\caption{Composition of the HSR finetuning dataset.}
\label{tab:hsr-dataset-composition}
\begin{tabular}{lr}
\toprule
\textbf{Task family} & \textbf{Trajectories} \\
\midrule
Mug Rectangle & 399 \\
Coffee Bottle $\rightarrow$ Box & 295 \\
Box Rearrangement & 195 \\
Coffee Bottles $\rightarrow$ Table & 316 \\
\bottomrule
\end{tabular}
\end{table}

\textbf{Coffee Bottle $\rightarrow$ Box} requires HSR to move from a start position to a coffee bottle, grasp it, navigate to the area in front of the designated box, and insert the bottle into that box. This task is difficult because the clearance above the target box is small, so locomotion error makes precise placement substantially harder.

\textbf{Coffee Bottles $\rightarrow$ Table} requires moving two coffee bottles from a shelf to the table and placing them near the mugs. This task combines locomotion with repeated pick-and-place under clutter, and requires stable transport from an elevated shelf to a lower placement surface.

\textbf{Box Rearrangement} requires moving the box labeled 2 beside the box labeled 1. Its subtasks are moving to box 2, grasping box 2, navigating to the area in front of box 1, and placing box 2 beside box 1. The box is large and relatively heavy for the gripper, making stable transport and final placement both challenging.

\textbf{Mug Rectangle} requires moving mugs so that the final layout forms a rectangular arrangement. The policy must reason about which mug should move and where the missing rectangle corner lies before executing the relocation.

\paragraph{Canonical HSR subtask sequences.}
\textbf{Mug Rectangle.} A typical sequence is: pick up the mug that is not at a rectangle corner, and place it at the missing rectangle corner.

\textbf{Coffee Bottle $\rightarrow$ Box.} A typical sequence is: pick up the coffee bottle on the right, and place it into the box labeled 1.

\textbf{Box Rearrangement.} A typical sequence is: pick up the box labeled 2, and place it beside the box labeled 1.

\textbf{Coffee Bottles $\rightarrow$ Table.} A typical sequence is: pick up the right-hand bottle from the shelves, place it next to any mug on the table, pick up the left-hand bottle from the shelves, and place it next to a mug that does not already have a bottle.

\begin{figure}[H]
\centering
\begin{minipage}[t]{0.48\textwidth}
    \centering
    \includegraphics[width=\textwidth]{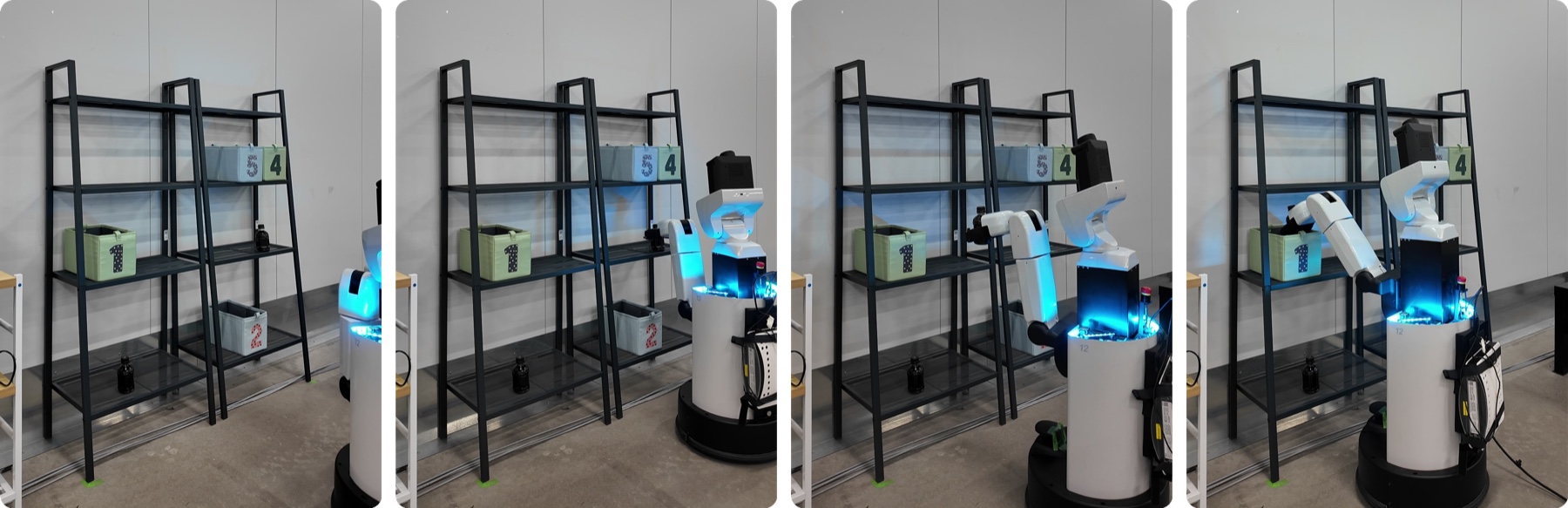}
    \par\vspace{0.2em}
    {\scriptsize \textbf{Coffee Bottle $\rightarrow$ Box}}
\end{minipage}\hfill
\begin{minipage}[t]{0.48\textwidth}
    \centering
    \includegraphics[width=\textwidth]{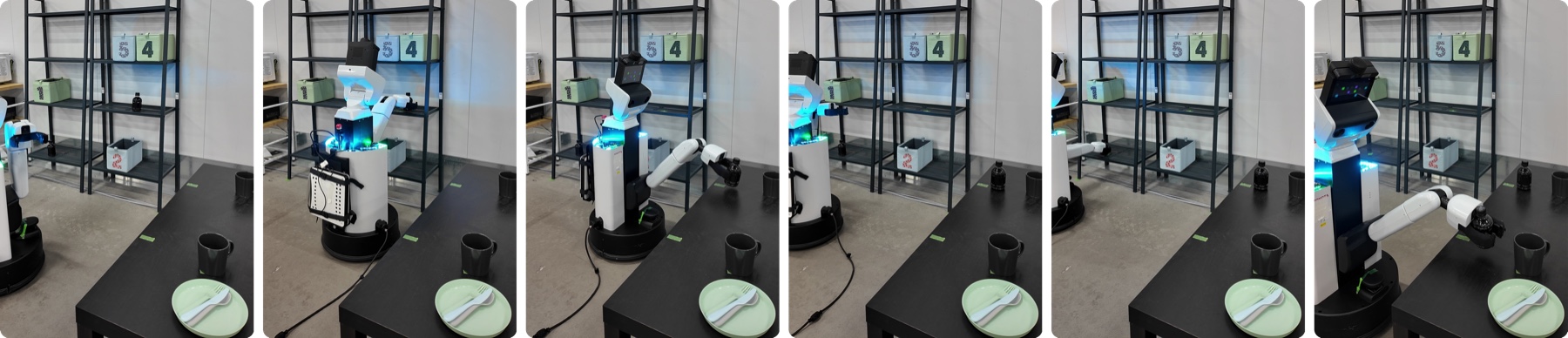}
    \par\vspace{0.2em}
    {\scriptsize \textbf{Coffee Bottles $\rightarrow$ Table}}
\end{minipage}

\vspace{0.6em}

\begin{minipage}[t]{0.48\textwidth}
    \centering
    \includegraphics[width=\textwidth]{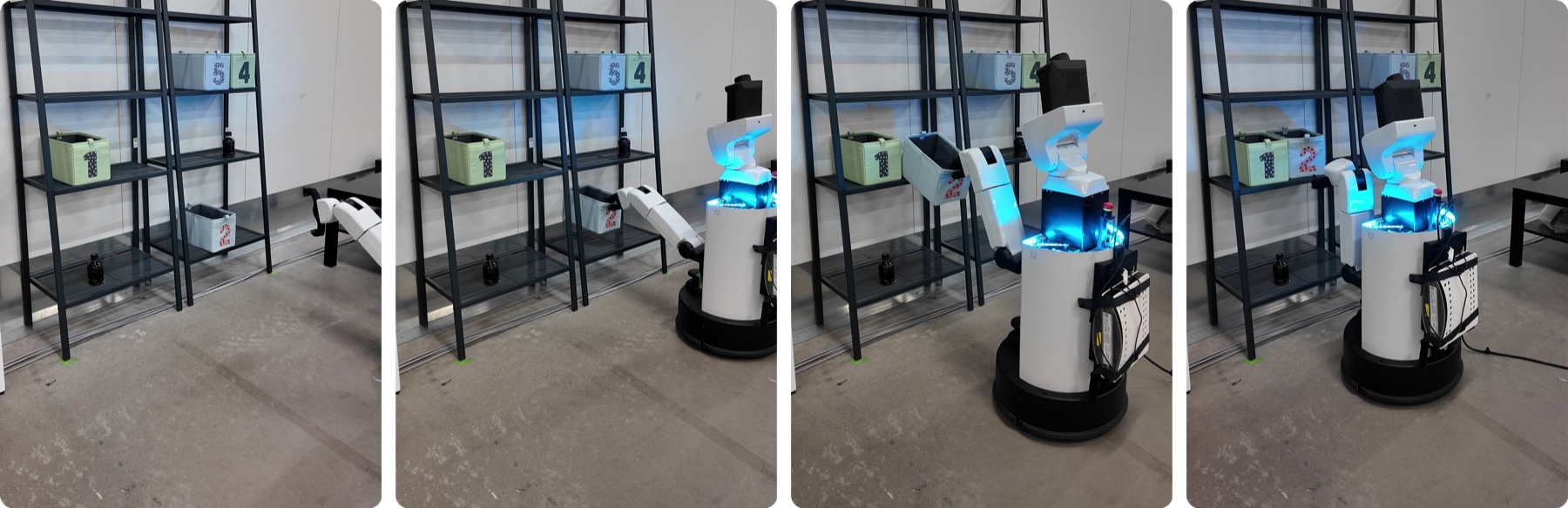}
    \par\vspace{0.2em}
    {\scriptsize \textbf{Box Rearrangement}}
\end{minipage}\hfill
\begin{minipage}[t]{0.48\textwidth}
    \centering
    \includegraphics[width=\textwidth]{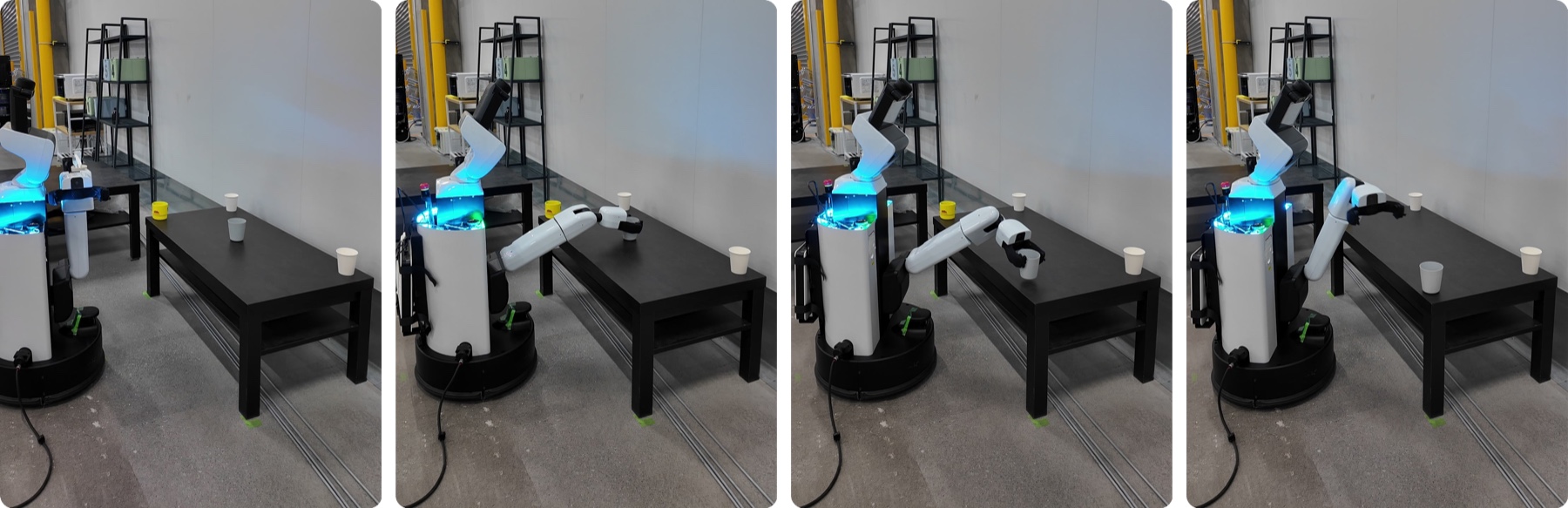}
    \par\vspace{0.2em}
    {\scriptsize \textbf{Mug Rectangle}}
\end{minipage}
\caption{Representative HSR trajectory snapshots for the four mobile-manipulation tasks described above. Each panel shows a rollout of the corresponding task family used in our real-robot evaluation.}
\label{fig:appendix-hsr-trajectories}
\end{figure}

\paragraph{Full real-robot tables.}
In the main-text real-robot table, TX-G2 shorthand headers \emph{Cutl.}, \emph{Bowl}, \emph{Cloth.}, and \emph{Dish} denote \emph{Cutlery Transfer}, \emph{Bowl Stacking}, \emph{Clothes Sorting}, and \emph{Dish Racking}, respectively. HSR shorthand headers \emph{Bott.-1}, \emph{Bott.-2}, \emph{Box}, and \emph{Mug} denote \emph{Coffee Bottle $\rightarrow$ Box}, \emph{Coffee Bottles $\rightarrow$ Table}, \emph{Box Rearrangement}, and \emph{Mug Rectangle}, respectively.

Table~\ref{tab:appendix-g2-real} reports the full TX-G2 metrics, including both mean subtask success and E2E completion. Table~\ref{tab:appendix-hsr-real} provides the corresponding HSR table with the same metrics.

\begin{table}[H]
\centering
\scriptsize
\setlength{\tabcolsep}{2.4pt}
\renewcommand{\arraystretch}{1.16}
\newcommand{\stdcell}[2]{\mbox{#1 {\color[gray]{0.45}\fontsize{5}{5}\selectfont$\pm$#2}}}
\caption{Full TX-G2 real-robot results on four long-horizon tasks. Subtask denotes mean subtask success with standard deviation, while E2E measures full-task completion when all subtasks succeed consecutively.}
\label{tab:appendix-g2-real}
\begin{tabular}{>{\raggedright\arraybackslash}m{2.00cm}|*{2}{>{\centering\arraybackslash}m{1.08cm}}|*{2}{>{\centering\arraybackslash}m{1.08cm}}|*{2}{>{\centering\arraybackslash}m{1.08cm}}|*{2}{>{\centering\arraybackslash}m{1.08cm}}}
\toprule
\multirow{2}{2.00cm}{\textbf{Method}} & \multicolumn{2}{c|}{\textbf{\shortstack{Cutlery\\Transfer}}} & \multicolumn{2}{c|}{\textbf{\shortstack{Bowl\\Stacking}}} & \multicolumn{2}{c|}{\textbf{\shortstack{Clothes\\Sorting}}} & \multicolumn{2}{c}{\textbf{\shortstack{Dish\\Racking}}} \\
& Subtask & E2E & Subtask & E2E & Subtask & E2E & Subtask & E2E \\
\midrule
\mbox{X-VLA} & \stdcell{0.0}{0.0} & 0.0 & \stdcell{7.5}{4.0} & 0.0 & \stdcell{5.0}{2.6} & 0.0 & \stdcell{5.0}{3.4} & 0.0 \\
\mbox{VLA-Adapter} & \stdcell{0.0}{0.0} & 0.0 & \stdcell{0.0}{0.0} & 0.0 & \stdcell{0.0}{0.0} & 0.0 & \stdcell{0.0}{0.0} & 0.0 \\
\pihalf{} & \stdcell{5.0}{3.2} & 0.0 & \stdcell{47.5}{6.8} & 0.0 & \stdcell{83.3}{4.8} & 50.0 & \stdcell{60.0}{7.2} & 20.0 \\
\rowcolor{TableHighlight}
CR (Ours) & \stdcell{\textbf{22.5}}{6.4} & \textbf{0.0} & \stdcell{\textbf{70.0}}{7.2} & \textbf{40.0} & \stdcell{\textbf{95.0}}{2.7} & \textbf{80.0} & \stdcell{\textbf{87.5}}{4.5} & \textbf{60.0} \\
\bottomrule
\end{tabular}
\end{table}
\global\let\stdcell\relax

\paragraph{Why does VLA-Adapter fail on TX-G2?}
We did not find evidence of a large inference-time integration error for VLA-Adapter. Its preprocessing path, proprioceptive normalization, and bundle-level input/output contract all passed our validation checks. However, in a 30-sample dataset-replay sanity test, the predicted actions were only marginally better than a simple mean-action baseline. This suggests that the adapted policy remained close to an average regressor rather than becoming a strongly task-conditioned controller, which is consistent with its 0\% closed-loop success on TX-G2.

\begin{table}[H]
\centering
\scriptsize
\setlength{\tabcolsep}{2.4pt}
\renewcommand{\arraystretch}{1.16}
\newcommand{\stdcell}[2]{\mbox{#1 {\color[gray]{0.45}\fontsize{5}{5}\selectfont$\pm$#2}}}
\caption{Full HSR real-robot results on four long-horizon mobile-manipulation tasks. Subtask denotes mean subtask success with standard deviation, while E2E measures full-task completion when all subtasks succeed consecutively.}
\label{tab:appendix-hsr-real}
\begin{tabular}{>{\raggedright\arraybackslash}m{1.90cm}|*{2}{>{\centering\arraybackslash}m{1.10cm}}|*{2}{>{\centering\arraybackslash}m{1.10cm}}|*{2}{>{\centering\arraybackslash}m{1.00cm}}|*{2}{>{\centering\arraybackslash}m{1.00cm}}}
\toprule
\multirow[c]{2}{1.90cm}{\textbf{Method}} & \multicolumn{2}{c|}{\textbf{\shortstack{Bottle\\Box}}} & \multicolumn{2}{c|}{\textbf{\shortstack{Two\\Bottles}}} & \multicolumn{2}{c|}{\textbf{\shortstack{Box\\Rearr.}}} & \multicolumn{2}{c}{\textbf{\shortstack{Mug\\Rect.}}} \\
& Subtask & E2E & Subtask & E2E & Subtask & E2E & Subtask & E2E \\
\midrule
X-VLA & \stdcell{8.3}{7.6} & 0.0 & \stdcell{4.2}{3.8} & 0.0 & \stdcell{0.0}{0.0} & 0.0 & \stdcell{25.0}{12.3} & 16.7 \\
VLA-Adapter & \stdcell{0.0}{0.0} & 0.0 & \stdcell{0.0}{0.0} & 0.0 & \stdcell{0.0}{0.0} & 0.0 & \stdcell{0.0}{0.0} & 0.0 \\
\pihalf{} & \stdcell{\textbf{83.3}}{9.6} & \textbf{66.7} & \stdcell{45.8}{8.5} & 0.0 & \stdcell{41.7}{14.0} & \textbf{16.7} & \stdcell{66.7}{9.6} & 33.3 \\
\rowcolor{TableHighlight}
CR (Ours) & \stdcell{\textbf{83.3}}{10.8} & \textbf{66.7} & \stdcell{\textbf{83.3}}{6.8} & \textbf{66.7} & \stdcell{\textbf{58.3}}{12.3} & \textbf{16.7} & \stdcell{\textbf{75.0}}{10.2} & \textbf{50.0} \\
\bottomrule
\end{tabular}
\end{table}
\global\let\stdcell\relax

\end{document}